\documentclass[runningheads]{llncs}

\usepackage[utf8]{inputenc}
\usepackage[T1]{fontenc}
\usepackage[hidelinks]{hyperref}
\usepackage{cleveref}
\usepackage{orcidlink}
\usepackage{subfig}
\usepackage{csquotes}
\usepackage{graphicx}
\usepackage{listings}

\PassOptionsToPackage{%
  backend=bibtex,
  language=auto,%
  style=numeric-comp,%
  sorting=anyt, %
  maxbibnames=20, %
  maxcitenames=2, %
  backref=false,%
  natbib=true, %
  urldate=iso8601,
  sortcites=false, %
  labelnumber,
  url=false
}{biblatex}
\usepackage{biblatex}

\begin{document}

\mainmatter

\title{Towards Self-organizing Personal Knowledge Assistants in Evolving Corporate Memories}

\author{
  Christian Jilek\inst{1,2}\orcidlink{0000-0002-5926-1673} \and
  Markus Schröder\inst{1}\orcidlink{0000-0001-8416-0535} \and
  Heiko Maus\inst{1}\orcidlink{0000-0003-3508-5860} \and
  Sven Schwarz\inst{1}\orcidlink{0009-0005-3064-5592} \and
  Andreas Dengel\inst{1,2}\orcidlink{0000-0002-6100-8255}
}

\authorrunning{C.~Jilek, M.~Schröder, H.~Maus, S.~Schwarz, and A.~Dengel}
\titlerunning{Towards Self-organizing PKA in Evolving Corporate Memories}

\institute{
  Smart Data and Knowledge Services Department, German Research Center for Artificial Intelligence (DFKI), Kaiserslautern, Germany \and
  Department of Computer Science, RPTU Kaiserslautern-Landau, Kaiserslautern, Germany\\
  \email{christian.jilek@dfki.de, markus.schroeder@dfki.de, heiko.maus@dfki.de,\\sven.schwarz@dfki.de, andreas.dengel@dfki.de}
}

\maketitle

\begin{abstract}
This paper presents a retrospective overview of a decade of research in our department towards self-organizing personal knowledge assistants in evolving corporate memories.
Our research is typically inspired by real-world problems and often conducted in interdisciplinary collaborations with research and industry partners.
We summarize past experiments and results comprising topics like various ways of knowledge graph construction in corporate and personal settings, Managed Forgetting and (Self-organizing) Context Spaces as a novel approach to Personal Information Management (PIM) and knowledge work support.
Past results are complemented by an overview of related work and some of our latest findings not published so far.
Last, we give an overview of our related industry use cases including a detailed look into CoMem, a Corporate Memory based on our presented research already in productive use and providing challenges for further research.
Many contributions are only first steps in new directions with still a lot of untapped potential, especially with regard to further increasing the automation in PIM and knowledge work support.


\end{abstract}

\section{Introduction}
\label{sec:Intro}
For more than 20 years now, the \emph{Smart Data \& Knowledge Services} department (formerly \emph{Knowledge Management} department) at the \emph{German Research Center for Artificial Intelligence (DFKI)} has investigated methods for supporting users' (personal) information management (PIM) \cite{Jones2008} and knowledge work \cite{Davenport2005} activities.
Assuming that knowledge\footnote{
  Regarding the definition of ``knowledge'', we adopt the \emph{knowledge ladder} by \citet{NorthBrandnerSteininger2016}:
In short, it starts with \emph{symbols} at the bottom. Adding \emph{syntax} leads to \emph{data}, adding \emph{meaning} leads to \emph{information} and \emph{connecting information} including personal context, experience and expectations leads to \emph{knowledge}.
} emerges with the individual and eventually spreads into groups like the team, the department or the company, our solutions are typically centered around an \emph{Organizational} or \emph{Corporate Memory}, i.e. a computer system that continuously collects, updates and structures knowledge and information in the organization and context-sensitively, specifically and actively provides them for various tasks (see \citet{AbeckerBernardiHinkelmann+98,Dengel2002} and \Cref{fig:omlayers}).
Especially in systems developed in our department, there is a \emph{Knowledge Graph (KG)} \cite{DBLP:journals/semweb/Paulheim17} based on ontologies \cite{StuderBenjaminsFensel1998} at its core to represent (parts of) users' mental models (like persons, organizations, locations, topics, tasks, events, etc.) and interconnect them with personal and corporate information (like files, mails, bookmarks, databases, etc.).
Based on such a Corporate Memory, \emph{Personal Knowledge Assistants (PKA)} operating on users' individual desktops as well as the corporate intranet (cloud service) can be implemented.

In interdisciplinary research with cognitive psychologists and ergonomists, our department has worked on several variants and aspects of such assistants over the years.
In this paper, we would like to provide an overview of our research of the last decade whose focus was in particular on further increasing the degree of automation in PIM and knowledge work support.
One major aspect in this regard was \emph{digital forgetting} and in particular \emph{Managed Forgetting} \cite{JilekRungeNiederee+2019} serving as a ``final puzzle piece'' to realize true \emph{self-organization} in this area:
Without the ability to also select information items to be archived, condensed, possibly deleted, a system -- as autonomous as it may be -- can more or less only ``move items around'' without ever ``shedding ballast'', even if indicators would advise to do so.
In order to decide which support measures to apply, the system observes user behavior and uses specifically tailored knowledge graphs for sense making and representing the user's information sphere.

The rest of this paper is structured as follows:
\Cref{sec:Background} introduces necessary background information and foundations.
An overview of related work is given in \Cref{sec:RelatedWork}.
Since evaluation can be very challenging in PIM and knowledge work scenarios,  \Cref{sec:Eval} addresses this matter in more detail presenting various challenges and solution strategies.
The paper's main part is split into three sections:
\Cref{sec:KGC} explains how we construct KGs from personal and corporate data, in particular often messy data.
Then, applications based on these KGs are presented:
\Cref{sec:PKA} details our research on self-organizing personal knowledge assistants, whereas industry use cases are showcased in \Cref{sec:Industry}.
Last, \Cref{sec:Conclusions} concludes the paper.

Note: a lot of the presented work has been conducted as part of the PhD theses by Markus Schröder (\emph{Building Knowledge Graphs from Messy Enterprise Data} \cite{Schroeder2022PhD}) and Christian Jilek (\emph{Self-organizing Context Spaces} \cite{Jilek2023PhD}).
Thus, for most sections of this paper a more detailed variant is available in the respective thesis.
\section{Background and Foundations}
\label{sec:Background}
This section introduces concepts and ideas that are key to our approach.

\subsection{Constructing Knowledge Graphs by Data Acquisition and Leveraging}
When creating Corporate Memory systems and PKAs based on them, we first need to solve the problem of \emph{data acquisition}, i.e. tapping into and connecting existing data sources on users' personal computers, the intra- and internet.
This data space can become heterogeneous, arbitrarily structured, diverse and is distributed in isolated stores \cite{DBLP:journals/sigmod/FranklinHM05}.
As a result, it shows a low data quality and is therefore not fit for its intended use \cite{DBLP:books/daglib/0093293}.
For example, we often come across short ungrammatical text snippets which are about a special domain, contain technical (or personal) terms and are not comprehensible without expert knowledge \cite{DBLP:conf/icde/HuaWWZZ15}.
A prominent example are file names which can mention technical terms, made-up words and even puns \cite{carroll1982creative}.
Due to various file naming strategies, words can be arbitrary concatenated and differently ordered \cite{DBLP:journals/tois/HicksDPM08,crowder2015file}.
On the lowest level, data acquisition is typically tackled by crawlers and connection adapters.
Next, a leveraging step is performed that adds semantics and interconnects items leading to a KG ``on top'' of the data.
Since sufficient documentation is usually not available (some knowledge is only available in persons' minds), it is necessary to consult domain experts making KG construction by data acquisition and leveraging an iterative, semi-automatic process.
For example, the system may come up with first results like extracted concepts, relations or contexts that are then assessed and refined by the user in a \emph{Human-in-the-Loop (HumL)} \cite{department2013dod} approach.
In follow-up iterations, the system can take such user feedback into account to improve its information extraction and leveraging results.
HumL models are typically found in \emph{interactive machine learning}, where humans actively participate in an algorithm's learning phase \cite{DBLP:journals/braininf/Holzinger16}.
However, in contrast to typical crowdsourcing scenarios, we usually have only a small number of experts and limited time available. 
Since assistance systems typically suffer from a so-called \emph{cold start problem}, i.e. the assistant not knowing anything about its users in the beginning (e.g. information items, observed interaction data, etc.), it is not able to support them well in its initial hours of usage.
Thus, data acquisition and leveraging are a crucial bootstrapping step in such scenarios.
Besides the bootstrapping, users of a PKA can rightly expect the assistant to consider and have available their personal as well as corporate and public data in its reasoning and support measures.

\subsection{The Semantic Desktop as an Ecosystem}
Such a personal or corporate KG is a core pillar of the Corporate Memory system and the PKA based on it.
Especially solutions developed in our department involve the \emph{Semantic Desktop} \cite{DeckerFrank2004,SauermannBernardiDengel2005} as an ecosystem bridging users' local devices with the Corporate Memory residing on the intranet.
Technically, the Semantic Desktop aims at bringing \emph{Semantic Web} \cite{BernersLeeHendlerLassila2001} technology to users' desktops.
In short, information items (files, emails, bookmarks, etc.) are treated as a Semantic Web resources, each identified by a \emph{URI} \cite{rfc2396,rfc3986} and accessible and queryable as an \emph{RDF} graph \cite{w3cRdf2004,w3cRdf2014}.
\emph{Ontologies} \cite{StuderBenjaminsFensel1998} allow users to express personal mental models that interconnect these information items with their mental concepts like persons, organizations, locations, projects, topics, tasks, events, etc.
And easy way offered by the Semantic Desktop is users simply tagging items with these concepts.
Such a resulting semantic network at the core of the Semantic Desktop is therefore also called the \emph{Personal Information Model} \cite{SauermannVanElstDengel2007}, or \emph{PIMO} for short.
As mentioned in the introduction, knowledge eventually spreads into groups, which is also reflected by the system: shared parts of users' individual PIMOs form a \emph{Group Information Model (GIMO)} \cite{MausSchwarzDengel2013} as part of the Corporate Memory.
In more recent terms, one may speak of PIMO and GIMO as a \emph{Personal KG} \cite{DBLP:conf/ictir/BalogK19} or \emph{Corporate/Enterprise KG} \cite{DBLP:conf/iceis/GalkinAVS17}, respectively.
Several features of our recent Semantic Desktop prototype that is part of our productively used Corporate Memory system, \emph{CoMem}, are illustrated in \citet{MausJilekSchwarz2018,JilekChwalekSchwarz+2019} and the CoMem website\footnote{
  \url{https://comem.ai/}
}.

Our department was among the Semantic Desktop pioneers bringing up the idea and technology in the early 2000s \cite{Sauermann2009PhD}.
Although the hype around the topic finally subsided about a decade later, our department kept working in this area continuously.
One of the more recent features using this technology as an ecosystem is \emph{Managed Forgetting}, a variant of intentional forgetting, discussed in the next section.

\subsection{Managed Forgetting -- A Variant of Intentional Forgetting}
According to \citet{DraganDecker2012} (2012), one of the reasons why the Semantic Desktop was still not widespread at the time, although being superior to conventional desktops as pointed out by \citet{FranzScherpStaab2009}, was the absence of a ``killer app'', i.e. a highly beneficial feature (possibly not to be found elsewhere) making users adopt the technology.
Managed Forgetting tries to fill this gap with its goal to increase the degree of automation of the Semantic Desktop in PIM and knowledge work support scenarios.

Managed Forgetting was envisioned in the EU project \emph{ForgetIT}\footnote{
  \url{https://www.forgetit-project.eu/}
} (2013--2016) that brought together an interdisciplinary team of eleven partners from different universities, research institutes and companies across Europe.
The idea was initially published by members of the team in two position papers \cite{KanhabuaNiedereeSiberski2013,NiedereeKanhabuaGallo+15} and later extended by us in the \emph{Managed Forgetting}\footnote{
  \url{http://www.spp1921.de/projekte/p4.html.en}
} project (2016--2023) emphasizing more on self-(re)organization to finally read as follows:\\
``Managed Forgetting is an approach performed by a computer system to replace the binary keep-or-delete paradigm with an escalating set of measures instead.
These measures range from 1) temporal hiding and inhibition, to 2) condensation, aggregation and summarization, to 3) adaptive reorganization, synchronization, archiving and deletion.
The selection and application of measures, i.e. determining what to forget and what to focus on, is performed in a self-organizing and decentralized way based on observed evidences (typically gathered by user activity tracking).
Managed Forgetting relies on two variants of information value assessment inspired by human memory and cognition:
\emph{Memory Buoyancy} for short- and medium-term information value and the \emph{Preservation Value} as a long-term counterpart.
The goal of Managed Forgetting is to complement not copy or replace human memory'' \cite{Jilek2023PhD}.

The definition especially mentions Memory Buoyancy and Preservation Value as two variants of \emph{Information Value Assessment}, i.e. measures trying to assign a score to information items reflecting their present or future value for the user.
For example, a train ticket may have a high Memory Buoyancy score while one is on the business trip, but its value rapidly decreases once the trip is over and the ticket cannot be used for another ride.
Although aiming for different perspectives (short/medium- vs. long-term), both concepts incorporate similar dimensions in their calculations like user activity and user investment, gravity, social aspects, etc. \cite{MausJilekSchwarz2018,JilekChwalekSchwarz+2019}.
In particular, Memory Buoyancy follows the metaphor that items losing relevance gradually ``sink away'', while those that are important (again) are pushed closer to the user by their higher buoyancy \cite{KanhabuaNiedereeSiberski2013,NiedereeKanhabuaGallo+15}.
Both concepts will be addressed in more detail in \Cref{sec:PKA}.

Another aspect worth pointing out is the implied paradigm shift allowing the system to reorganize items on a person's computer and especially select some of them for deletion (as well as executing the deletion at some point in time).
Since the time computers became \emph{personal} computers (i.e. a great number of people having one or more devices for themselves), users could rely on information being stored on their computer would remain there (hardware malfunction or manual deletion aside).
This changes when introducing self-organizing and especially \emph{Forgetting-enabled (``forgetful'') Information Systems (FIS)} \cite{JilekRungeNiederee+2019} imposing new challenges to usability, users' feeling of control, trust etc.
Imagine a user entering keywords into the search field of an FIS and no (or seemingly incomplete) results are shown.
Users would wonder whether they entered the ``right'' keywords or whether they really saved the information item they are now looking for.
Searching FIS as well as the aforementioned challenges are addressed more thoroughly in \Cref{sec:PKA}.

In the Managed Forgetting approach, deciding, for example, which items to forget and which to focus on heavily relies on \emph{context}: the same information item could be very important in one context while being totally irrelevant in another.
Thus, another rather recent feature based on Semantic Desktop technology we envisioned are \emph{Context Spaces} \cite{JilekSchroederSchwarz+2018}.
They are introduced in the next section.

\subsection{Context Spaces -- User Context as an Explicit Interaction Element}
When speaking of ``context'' we actually mean ``a `sense-giving environment' for a (given) nucleus, i.e. a context tries to represent relevant information items and their relations describing the given situation.
Such a nucleus can be an activity (e.g. writing a scientific paper), an event (a meeting) or an information item itself
(a PDF document, email, etc.).
Because of the dynamics of situations, a context evolves over time.
The context of a large research task (later containing many documents, links, notes, etc.), for example, could spawn from a small context having only an email calling for participation as its nucleus'' \cite{GauselmannRungeJilek+2022}.
Since every information item could become the nucleus of such a context, this definition leaves great freedom to the user.
What belongs together in a person's mind, should be representable as a context (or \emph{Context Space} \cite{JilekSchroederSchwarz+2018}) on their computer.
In one of our industry use cases (see \Cref{sec:Industry}), there was a context space with a ticket as its nucleus for every failure incident reported by a customer or system health monitor.
Like in folders, items in the same context should then also be displayed together, so their relatedness is clearly visible to the user countering the well-known \emph{project fragmentation problem in PIM} \cite{Bergman2006}, i.e. items of the same context being spread across various applications (files in the file system, mails in the email client, bookmarks in the web browser, etc.).
This latter aspect is addressed by injecting context spaces into as many applications as possible.
Details will be presented in \Cref{sec:PKA}.

In summary, with context spaces we see context as an active element users can work \emph{in} and interact \emph{with}.
There are similarities to traditional folders but the idea of context spaces extends this concept as explained in the next section.

Having introduced these concepts, the next section presents \emph{Self-organizing Context Spaces} \cite{Jilek2023PhD} as a variant of a PKA that combines these ideas.

\subsection{Self-organizing Context Spaces as a PKA Variant}
The idea of Self-organizing Context Spaces, or \emph{cSpaces} for short, is to combine Semantic Desktop technology with Managed Forgetting and Context Spaces in the following way \cite{Jilek2023PhD}:
\begin{itemize}
\item Have context as an explicit interaction element: \emph{Context Spaces} allow knowledge workers to work \emph{with} (i.e.
a ``tangible'' object similar to a folder) and \emph{in} contexts (i.e. immersion).
These Context Spaces should be available in as many applications as possible fulfilling the vision of a single unified hierarchy structure (e.g. a tree or a directed acyclic graph) replacing separate file, mail, web browser and similar folder structures.
Additionally, they provide necessary contextual information to select and perform appropriate support measures.\\

\item Apply measures of \emph{Managed Forgetting} like temporal hiding, condensation, reorganization, etc. to support knowledge workers with new kinds of services and higher degrees of automation not available in traditional systems.
Exploiting the high amount of contextual (meta-)information makes this possible.
Capabilities like condensation, summarization and ultimately deletion may significantly help with regard to long-term scalability of such a system (another reason for failure of preceding Semantic Desktop prototypes).\\

\item Use the \emph{Semantic Desktop} as an ecosystem to capture and represent a user's personal information model (PIMO) as well as contextual meta-information.
The latter is either gathered automatically by means of user activity tracking or manually by users working with Semantic Desktop features like tagging or dragging items to a certain Context Space.
An enhanced PIMO explicitly contains all of a user's Context Spaces.
\end{itemize}

As mentioned in the introduction, Managed Forgetting is not seen in isolation but as a component to realize self-organization.
Thus, cSpaces aims at supporting the whole lifecycle of a context, from its initial spawning, to potential actions like splitting or temporal hiding in its main phase, to merging, condensation and potentially forgetting, archiving or deletion in its late phase.
More details on these operations are provided in \citet{Jilek2023PhD} and a selection is also presented in \Cref{sec:PKA}.

Note that cSpaces as a whole does not qualify as a self-organizing system according to the definition by Heylighen and Gershenson \cite{HeylighenGershenson2003}.
This has not been claimed, the term rather refers to the way cSpaces maintains its content (i.e. context spaces and their items) in a self-organizing and decentralized way by using operations like the ones described before (also see Managed Forgetting definition above).

The whole concept of Context Spaces and especially Self-organizing Context Spaces is explained in great detail in \citet{Jilek2023PhD}.\\

\begin{figure}
  \centering
  \includegraphics[width=1\columnwidth]{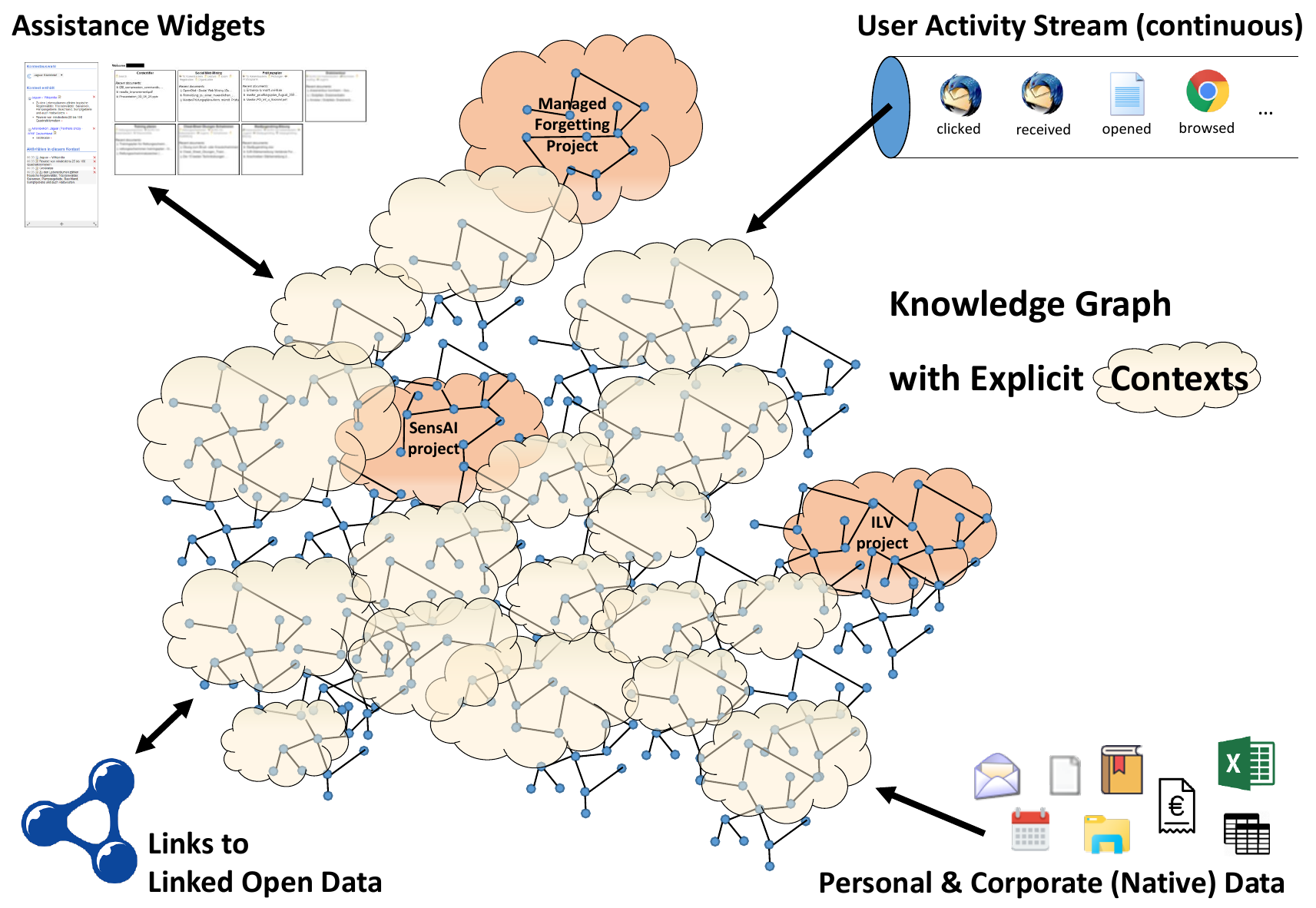}
  \captionsetup{format=hang}
  \caption{Technical scenario}
  \label{fig:cSpacesScenario}
\end{figure}

Concluding this section, \Cref{fig:cSpacesScenario} shows an overview of our technical scenario:
In the figure's middle is the KG with explicit context spaces (drawn as clouds; three of them exemplary highlighted in a different color).
The KG is connected (refers) to personal and corporate data (bottom right) and can be linked to \emph{Linked Open Data} \cite{BernersLee2006} sources like \emph{DBpedia}\footnote{
  \url{https://www.dbpedia.org/}
} or \emph{Wikidata}\footnote{
  \url{https://www.wikidata.org/}
} (bottom left).
Contexts are associated with evidence snippets (opened files, clicked webpages, etc.) coming from a continuous user activity event stream (upper right).
Last, there is the user interface allowing to interact with various assistance widgets (upper left).
For further technical details please see \citet{Jilek2023PhD}.

\section{Related Work}
\label{sec:RelatedWork}
Since this is an overview paper, a lot of related work could be cited here.
However, we restrict ourselves to a high-level overview and kindly refer the reader to our individual papers on a particular topic for further pointers to related work.

\subsection{Knowledge Graph Construction}
The construction of knowledge graphs (KGs) from input data as mentioned before has the goal to formally organize their contents as entities interrelated in graphs \cite{DBLP:journals/semweb/Paulheim17}.
This way, they serve as semantic bridges between domain conceptualization and raw data \cite{Schroeder2019}.
The focus usually lies on instances (i.e. persons, projects, topics) rather than on their terminology (classes and properties).
To acquire KGs, a typical process (or lifecycle) model involves knowledge acquisition, their curation (i.e. assessment, cleaning, enrichment), and deployment \cite{DBLP:books/sp/FenselSAHKPTUW20}.
These process steps can be realized with fully-automatic, semi-automatic or manual approaches.
Methods include the interpretation of tables \cite{DBLP:journals/widm/BonfittoCM21}, semantic labeling \cite{DBLP:conf/semweb/PhamAKS16} and data lifting \cite{DBLP:conf/mtsr/FiorelliS20}.
The latter usually involves a mapping language to define how the knowledge graph should be constructed; a prominent one is the \emph{RDF Mapping Language (RML)} family and its extensions \cite{DBLP:conf/www/DimouSCVMW14,DBLP:conf/esws/MeesterDVM16,DBLP:conf/esws/DelvaAHMD21,DBLP:conf/semweb/DelvaAICCD21}.
RML engines \cite{DBLP:conf/esws/SimsekKF19,DBLP:conf/cikm/IglesiasJCCV20} are usually able to build KGs from XML, JSON, CSV files and relational databases.
For other data formats, like spreadsheets, several dedicated data lifting approaches can be found in literature \cite{DBLP:conf/semweb/HanFPSJ08,DBLP:conf/ieaaie/FiorelliLPST15,DBLP:conf/semweb/OConnorHM10,DBLP:conf/semweb/LangeggerW09,DBLP:conf/esws/Hammar20,DBLP:journals/jidm/BernardoMS13}.

Once humans are involved in forming a KG, approaches are usually associated with the field of \emph{Human-Computer Interaction (HCI)} \cite{HCI}.
There are various possibilities to acquire knowledge from participants, such as \emph{data collection} \cite{olson1987extracting,weller1988systematic}, \emph{document annotation} \cite{DBLP:conf/his2/ShengW000XLSZ19,DBLP:conf/semweb/GentileGRW19}, \emph{microtask completion} \cite{DBLP:journals/ws/AcostaSFV17}, \emph{data alignment} \cite{DBLP:conf/esws/ClarksonGGRTW18,DBLP:conf/ekaw/ElstK04} or \emph{data preprocessing} \cite{DBLP:conf/icadl/OelenSA20}.
\emph{Games with a purpose} \cite{DBLP:journals/cacm/AhnD08} are often utilized to curate KGs, for example, with crossword puzzles \cite{JovanovicGamifyingKnowledgeMaintenance}, rating \cite{hees2011betterrelations} or question answering \cite{DBLP:conf/aciids/BuK20}.
To let participants see and inspect the modeled KG, a common method is \emph{graph visualization and browsing} \cite{DBLP:conf/fruct/AntoniazziV18}.

\subsection{Semantic Desktop and Similar Approaches}
In general, there is the inspiring vision of \emph{Memex} by Vannevar Bush \cite{Bush1945} for many Semantic Desktop and similar approaches.

As mentioned before, our department has quite a long track record in the area of organizational/corporate memories in general and the Semantic Desktop in particular.
Some of the past projects to be mentioned are \emph{KnowMore}\footnote{
  \url{https://www.dfki.uni-kl.de/frodo/knowmore.html}
} (1997--1999), \emph{FRODO}\footnote{
  \url{https://www.dfki.uni-kl.de/frodo/}
} (2000--2002), \emph{EPOS}\footnote{
  \url{https://www.dfki.uni-kl.de/epos/}
} (2003--2005), \emph{Mymory}\footnote{
  \url{https://web.archive.org/web/20161114142751/http://www.dfki.uni-kl.de/mymory}
} (2006--2008), \emph{NEPOMUK}\footnote{
  \url{https://nepomuk.semanticdesktop.org/}
} (2006--2008) and \emph{ADiWa}\footnote{
  \url{https://web.archive.org/web/20150202200550/http://adiwa.net/}
} (2009--2012).
Especially NEPOMUK was a large EU project dedicated to the Semantic Desktop that had several spiritual successor projects like the aforementioned \emph{ForgetIT}\footnote{
  \url{https://www.forgetit-project.eu/}
} (2013--2016) and \emph{Managed Forgetting}\footnote{
  \url{http://www.spp1921.de/projekte/p4.html.en}
} (2016--2023) as well as \emph{supSpaces}\footnote{
  \url{https://web.archive.org/web/20190109230722/http://www.supspaces.de/}
} (2015--2017) and \emph{SensAI}\footnote{
  \url{https://comem.ai/sensai/}
} (2020--2023).
Relevant papers and software assets of these projects are far too many to be listed here.
Some of the highlights (judged by either their impact or relevance to this paper) are presumably the \emph{NEPOMUK Semantic Desktop Prototype}\footnote{
  \url{https://web.archive.org/web/20150920022727/http://dev.nepomuk.semanticdesktop.org/download/}
} and \emph{Nepomuk-KDE}\footnote{
  \url{https://web.archive.org/web/20130401085641/https://nepomuk.kde.org/}
}, first evaluations on the Semantic Desktop \cite{SauermannHeim2008,FranzScherpStaab2009}, the PhD theses by \citet{Maus2007PhD}, \citet{Sauermann2009PhD}, \citet{Schwarz2010PhD} and \citet{Schmidt2013PhD} (which also serve as a good overview of their individual papers) as well as \emph{CoMem}\footnote{
  \url{https://www.comem.ai/}
}, our Corporate Memory prototype.
For a more detailed but still concise overview please see \citet{Jilek2023PhD}.
Parts of the outcome of several later projects like ForgetIT, Managed Forgetting, supSpaces and SensAI are actually part of this paper (see Sections \ref{sec:Eval} to \ref{sec:Industry}).

Researchers of our department were involved in the works mentioned so far.
Apart from that, there is also work by Microsoft Research: \emph{MyLifeBits} \cite{GemmellBellLueder+2002,GemmellBellLueder+2006}, \emph{Stuff I've Seen} \cite{DumaisCutrellCadiz+2003}, the announced but later canceled \cite{WinFS2} \emph{Windows Future Storage (WinFS)} \cite{WinFS1}, the idea of \emph{Memory Landmarks} \cite{RingelCutrellDumais+2003,HorvitzDumaisKoch2004}, the \emph{Office Assistant} also known as \emph{Clippit} or \emph{Clippy} based on the \emph{Lumière Project} \cite{HorvitzBreeseHeckerman1998} and the assistant \emph{Cortana}\footnote{
  \url{https://www.microsoft.com/en-us/cortana/}
}.
There was the \emph{CALO} project (Cognitive Assistant that Learns and Organizes) with Semantic Desktops like \emph{SEMEX} \cite{DongHalevyNemes+2004} or \emph{IRIS} \cite{CheyerParkGiuli2005}.
The well-known \emph{Enron Email Dataset} \cite{KlimtYang2004} was also collected and prepared in this project.
However, its most prominent outcome is presumably the assistant \emph{Siri}\footnote{
  \url{https://www.apple.com/siri/}
} that started as a spin-off later bought by Apple\footnote{
  \url{https://web.archive.org/web/20220308041153/http://www.ai.sri.com/project/CALO}
}.
Besides Cortana and Siri, there are also \emph{Amazon Alexa}\footnote{
  \url{https://developer.amazon.com/alexa}
} and \emph{Google Assistant}\footnote{
  \url{https://assistant.google.com/}
} (formerly \emph{Google Now}) as further assistants by major corporations.
Google also provided the 2011 discontinued \emph{Google Desktop}\footnote{
  \url{https://web.archive.org/web/20110824073151/http://desktop.google.com:80/}
}.
Other works are \emph{TaskTracer} \cite{DragunovDietterichJohnsrude+2005,StumpfBaoDragunov+2005,ShenLiDietterich+2006,ShenLiDietterich2007} by Dietterich, Shen, Stumpf \emph{et al.} or papers on user activity tracking and task detection \cite{GyllstromSoulesVeitch2008,GyllstromStotts2006} by Gyllstrom \emph{et al.}, who also published the information value assessment approach \emph{LostRank} \cite{GyllstromPedersen2010} (see next section).
The group of Abela, Staff, Handschuh, Scerri \emph{et al.} (some of whom were also members of the NEPOMUK project) worked on several related topics and presented solutions like \emph{DCON} \cite{ScerriAttardRivera+2012}, \emph{PiMx(T)} \cite{AbelaStaffHandschuh2015} or \emph{(X-)iDeTaCt} \cite{AbelaStaffHandschuh2015b,AbelaStaff2016}.
Last, there were three large projects, \emph{APOSDLE}\footnote{
  \url{https://web.archive.org/web/20190503192712/http://www.aposdle.tugraz.at/}
} (2006--2010) \cite{LindstaedtMayer2006}, \emph{ACTIVE}\footnote{
  \url{https://web.archive.org/web/20110305113841/http://active-project.eu/}
} (2008--2011) \cite{WarrenThurlowDavies+2010} and \emph{SWELL}\footnote{
  \url{https://web.archive.org/web/20210612234945/http://swell-project.net/}
} (2011--2017) \cite{KraaijVerberneKoldijk+2020}, each publishing 60 or more papers, several of them relevant.
For a slightly more detailed overview see \citet{Jilek2023PhD}.

\subsection{Digital Forgetting and especially Managed Forgetting}
With regard to digital forgetting, presumably the most prominent topic known to a general audience is the \emph{Right to Be Forgotten}, which is codified in Europe in Article 17 of the \emph{General Data Protection Regulation (GDPR)} \cite{GDPR} and came into effect in 2016.
The idea actually originated from one of our colleagues in the ForgetIT project, Victor Mayer-Schönberger, who proposed expiration dates for information items after which they should automatically be deleted (if not explicitly renewed by the user) \cite{MayerSchonberger2009}.
Apart from this legal concept, there is a lot of other related work, to a large part published by our project partners.
In \citet{MezarisNiedereeLogie2018}, all eleven partners of the ForgetIT project summarized their contributions of three years of joint research.
The Managed Forgetting project was embedded in a special priority program (SPP 1921\footnote{
  \url{http://www.spp1921.de/}
}) by the German Research Foundation (DFG) spanning more than six years and bringing together eight interdisciplinary research teams on the topic of intentional forgetting in organizations.
Beside each project's individual publications (e.g. \cite{SiebersSchmid2019,SiebersGoebelNiessen+2017,GoebelNiessenNandini+2021,EiterKernIsberner2019}), the consortium also published several cross-project papers on their different perspectives on the topic as well as commonalities.
Some papers were written from a computer science perspective (e.g. \cite{TimmStaabSiebers+2018}), others more from the point-of-view of cognitive science \cite{EllwartUlfertAntoni+2019,EllwartKluge2019}.

Other works on digital forgetting are, for example, in the area of \emph{Robotic Memories} \cite{GurrinLeeHayes2010} or \emph{Life Logging} \cite{GurrinSmeatonDoherty2014}.
With regard to PIM, \citet{Schmidt2006} applied an ``aging mechanism as a form of controlled forgetting'' in user context management and \citet{BergmanBeythMaromNachmias2008} proposed a \emph{GrayArea} as ``an additional folder feature that allows the users to drag information items of low importance to a designated location at the bottom of a folder''.
There is also the area of \emph{Machine Unlearning}, i.e. the removal of individual data points from a trained model (e.g. \cite{CaoYang2015,BourtouleChandrasekaranChoquetteChoo+2021,GuptaJungNeel+2021}).

Regarding Information Value Assessment (see the introduced Memory Buoyancy and Preservation Value), related approaches are, for example, the ideas of \emph{waste data} \cite{HasanBurns2013} or \emph{information waste} in the file system \cite{WijnhovenAmritDietz2014} or on the internet \cite{AmritWijnhovenBeckers2015}.
Turczyk et al. \cite{Turczyk2009PhD} investigated \emph{file valuation} in the area of \emph{Information Lifecycle Management (ILM)}, which seeks to store files on different storage systems according to their (business) value.
\citet{GyllstromPedersen2010} proposed \emph{LostRank}, an approach to estimate which documents are most likely to be lost for the user (e.g. important but not recently used) and are thus harder to re-find.
\citet{SappelliVerberneKraaij2013} presented an approach for email importance estimation.
Last, there is work by Attard, Brennan et al. (e.g. \cite{AttardBrennan2018b,AttardBrennan2019,BrennanAttardPetkov+2019}) using the mostly synonymous term of \emph{data value assessment}.
Some of these approaches are more business-centric, others aim at more personal use cases.
However, to the authors' best knowledge, Memory Buoyancy and Preservation Value are the only approaches that were designed and implemented according to findings of cognitive psychology about human memory and cognition.

\subsection{Context Spaces and in particular Self-organizing Context Spaces}
Apart from our own prior research, especially work of the aforementioned ACTIVE project and the area of \emph{Semantic File Systems} are to some extent related to the idea of working with and in Context Spaces and Self-organizing Context Spaces.

ACTIVE assumed that users are aware of the concept of context \cite{GomezPerez2009} and allowed them to select their current one.
As a support measure, recently used document lists were updated according to the selected context \cite{GomezPerez2009,WarrenThurlowDavies+2010}.

Allowing users to actually work in Context Spaces ultimately requires making them available in many (ideally all) of the users' applications.
Especially the file system is interesting in this regard since it is available in all applications.
Thus, it is an effective approach to utilize virtual or semantic file systems.
To the authors' best knowledge, the paper that first mentioned the idea of a semantic file system is the one by \citet{GiffordJouvelotSheldon+1991} (1991).
Since then, more than 700 publications cited the paper, whereas only a fraction of them presents a semantic file system as well.
However, seeing a user's knowledge work context as an entity to traverse has not been realized by any of these systems.
The same is true for self-organization: none of these systems is self-(re)organizing except the one by \citet{MesnierThereskaGanger+2004}, which addresses and applies self-reorganization in the scenario of low-level file system organization but not with respect to users' knowledge work contexts.\\

Before coming to this paper's main chapters presenting various experiments and their results, the next section first gives some insights into evaluation challenges and strategies in typical PIM and knowledge work support scenarios.
Since experiments in this area can be quite challenging, we decided to dedicate a separate section to the matter.

\section{Evaluation Challenges and Strategies}
\label{sec:Eval}
In the research area of PIM and knowledge work support, evaluations can become tricky and challenging.
Therefore, this section gives an overview of challenges and possible solutions.
For us, these challenges resulted in a multi-lane evaluation strategy that is described in the second half of the section.

\subsection{Challenges and Possible Solutions}
We faced four major challenges in our experiments.
In the following, they are explained in more detail complemented by hints how we tackled them \cite{Jilek2023PhD}:

\begin{itemize}
\item \textbf{Subjectiveness}:
Users have subjective views on their data \cite{Dengel2006,BergmanBeythMaromNachmias2008}.
How could participants judge whether PIM and knowledge work support measures like forgetting or reorganization are correct and helpful if they never knew the involved information items?
Thus, in our experiments, we often try to enable participants to work on their own data.
This induces additional effort to first bootstrap the system with their data -- see PIM Crawler, concept and context mining in Section \ref{sec:KGC}.
In some experiments, like the photo preservation study mentioned in Section \ref{sec:PKA}, the system was bootstrapped by the experimenter conducting an initial interview with each participant and manually entering and creating relevant persons, locations, topics, etc. in the system.
These measures also counter the cold start problem mentioned in the beginning.\\

\item \textbf{Privacy issues}:
Systems collecting and storing potentially very sensitive user data should be privacy-protecting by design.
A study in the SWELL project about privacy and user trust in context-aware systems found that ``privacy information had a positive effect on perceived privacy and trust in [the] system'' \cite{KoldijkKootNeerincx+2014}.
This privacy information involved informing participants about 1) purpose limitation, 2) control, 3) data minimization, 4) data aggregation, 5) adequate protection and 6) data subjects right \cite{KoldijkKootNeerincx+2014}.
\citet{KoldijkKootNeerincx+2014} also found that the ``attitude towards using [the] system was related to personal motivation, and not related to perceived privacy and trust'', which is why they recommend to ``implement privacy by design to adequately protect the privacy of the users'' in such systems.

These findings are in line with measures we took in our experiments.
The following is a selection:
1) allowing users to temporarily disable user activity tracking, 2) only storing sensitive data on the user's local device without uploading anything to a server or cloud service, 3) only stimulating the semantic graph (e.g. ``activation'' of respective parts) without storing details permanently, 4) allowing users to view and delete tracked data at any time (especially if they are willing to agree to an anonymized submission of their data at the end of an experiment).
Individual measures may depend on the situation and the experiment's design: For example, items 1) and 2) were especially relevant for us in long-term studies.
With regard to the former, turning off the observation in a short-term laboratory study, for example, would practically mean quitting the experiment.

In our most recent user study that involved seven participants for a duration of five months, participants were asked whether they felt comfortable with our privacy preservation measures, to which six of seven \emph{quite} or \emph{strongly agreed}.\\

\item \textbf{Passing of time needed}:
Short-term experiments of only a few minutes (30 to 90) are often too short to get participants or the systems in a situation, in which several of our envisioned support measures would actually be applied.
Although some aspects can be simulated, testing a long-term measure by testing its modified variant for short-term experiments remains at least difficult due to possibly falsifying assumptions and modifications to be made.
An example, in which such a simulation worked quite well was seeing own data again after a long or very long time (only faint or no memories left), which is similar to seeing unknown data for the first time.
Nevertheless, the problem of needing time to pass led to a mixture of short- and long-term studies in our case.\\

\item \textbf{Missing datasets}:
To the authors' best knowledge, there is still no publicly available PIM and/or knowledge work dataset containing user contexts and all content browsed or worked with during the study.
Major obstacles are typically privacy and copyright issues.
\citet{Gonccalves2011} additionally argues that if such a dataset was available, it would still lack the semantic information to really make use of the data, e.g. whether a term is the name of a project or whether a mentioned person is a co-worker or spouse, etc.
Other approaches like Kim and Croft \cite{KimCroft2009a,KimCroft2009b} created \emph{pseudo desktop collections} for their experiments on information retrieval.
However, these collections neglect important sources like bookmarks or calendar events as well as structures like the file folder hierarchy, which also carry a lot of semantics.
For us, the most promising datasets are the previously mentioned \emph{Enron Email Dataset} and another one collected in the SWELL project.
Nevertheless, Enron's focus on emails only covers a fraction of PIM activities and data, and it neglects user contexts completely.
Regarding the latter, the \emph{SWELL Knowledge Work Dataset for Stress and User Modeling Research} \cite{KoldijkSappelliVerberne2014} contains user contexts but unfortunately lacks a lot of information items participants worked with during their study.
This is due to the aforementioned copyright reasons as stated by the authors themselves\footnote{
  \url{https://doi.org/10.17026/dans-x55-69zp}
}.

We tackled the problem of a missing publicly available dataset by enriching available ones to get closer to our scenarios (if possible) and conducting own data collections.
\end{itemize}

\subsection{Multi-lane Evaluation Strategy}
Tackling the challenges mentioned before resulted in a multi-lane evaluation strategy for us.
Given the characteristics of the experiment, time and funding constraints, we typically chose one of the following options \cite{Jilek2023PhD}:

\begin{itemize}
\item \textbf{Short-term studies}:
A first lane are short-term studies, preferably conducted under controlled laboratory conditions.
Several of our studies were conducted together with colleagues of cognitive psychology and ergonomy (e.g. \cite{GauselmannRungeJilek+2022,JilekGauselmannChwalek+2021}).
They typically involved a comparably large number of participants (about 20 to 50) but the exposure to the system was quite short (30 to 90 minutes).
As a consequence, some aspects needed to be ``artificially simulated'' or enforced to happen during the experiment (as stated earlier).\\

\item \textbf{Medium- or long-term studies}:
A second lane are medium- or long-term studies.
Our studies in this regard ranged from several weeks up to multiple years.
Mainly for cost reasons, less participants are typically involved in such studies but effects to be observed occur naturally due to the longer times of exposure to the system.
In our experiments, timespans of a few weeks typically still involved ``regular'' participants (i.e. persons unknown to the experimenter), whereas studies lasting several months or even years were usually conducted with fellow researchers or industry partners willing to use an experimental application for such a long time.

Publishing the full data obtained in such studies (especially tracked user activities and viewed content) is however a problem due to the aforementioned copyright and privacy issues as well as non-disclosure agreements with industry partners.
Anonymization and obfuscation (i.e. making content unintelligible for others but preserving certain features like word frequencies) could be an option.
However, these methods harbor the risk to be vulnerable to de-anonymization or de-obfuscation attacks, respectively.\\

\item \textbf{Inquiries}:
Another lane is to provide research and demo material to audiences and letting them perform a thought experiment like assuming a certain role in a hypothetical situation.
In the process or afterwards, they give their thoughts in an inquiry.
Such scenarios can comparably easily and cheaply be conducted with a large number of participants.
For example, our most recent online inquiry involved 140 participants \cite{Jilek2023PhD}.\\

\item \textbf{Data-driven studies}:
If datasets for a certain topic are available or similar datasets can be enriched or otherwise modified to fit another scenario, a completely data-driven approach can be followed.
For example, we evaluated a specific information extraction method using Wikipedia\footnote{
  \url{https://www.wikipedia.org/}
} articles \cite{JilekSchroederNovik+2019}.
In general, open government data and publicly available company data \cite{DBLP:conf/ecml/KlimtY04} might also be usable but their content rarely reflects PIM structures.\\

\item \textbf{Dataset generation}:
Data generation is a popular method when systems need to be tested \cite{DBLP:conf/icce-berlin/PopicPVT19}.
Various languages are proposed in literature to define how datasets, mainly relational databases, are generated \cite{DBLP:conf/vldb/BrunoC05,DBLP:journals/sigmod/HoagT07,DBLP:conf/kdd/JeskeSLYCXYLHR05,DBLP:conf/tpctc/RablFSK10,DBLP:conf/sigmod/RablP11}.

To generate documents and KGs for PIM scenarios, various contributions have been made by us.
A spreadsheet generator was proposed in \citet{SchroederJilekDengel2021patterns}, which utilizes messy patterns\footnote{
  \url{https://www.dfki.uni-kl.de/~mschroeder/pattern-language-spreadsheets/}
} to generate data and its KG.
Since personal information often involves mentioned person names in different variations, we developed a person index and text generator in \citet{SchroederJilekSchulze+2021}.
In \citet{SchulzeSchroederJilek+2021b}, a dedicated generator for invoices for the domain of purchase-to-pay \cite{TrkmanMcCormack2010} was designed, which additionally produces the expected KG using a suitable terminology \cite{SchulzeSchroederJilek+2021}.
Last, we created a generator for photo metadata like mentioned entities in a comment, connections to other entities in the KG, usage statistics, etc. -- more details will be given in Section \ref{sec:PKA} when presenting our photo preservation study.\\

\item \textbf{Simulation}:
The generators mentioned before only cover a very specific aspect.
A topic we intend to explore in the future, however, is trying to simulate \emph{comprehensive} PIM and knowledge work activities, especially to such an extent that generated and real data cannot be distinguished for one another anymore.
Since this would involve several agents interacting with each other, we envision a solution based on multi-agent systems \cite{DoranFranklinJennings1997}.

\end{itemize}

Having discussed several evaluation challenges in our research area, the next sections present experiments and results with regard to knowledge graph construction (Sec. \ref{sec:KGC}) and self-organizing personal knowledge assistants (Sec. \ref{sec:PKA}) followed by a section on industry use cases related to both topics (Sec. \ref{sec:Industry}).

\section{Knowledge Graph Construction}
\label{sec:KGC}

Knowledge Graph (KG) construction happens for us on two levels:
on a corporate and on a personal one.
While the former requires to analyze data managed on intranet or cloud storages, the latter involves the personal devices of employees which is usually a personal computer with a desktop system.
Regardless of the location, both provide useful but also messy data to construct KGs from.
The PhD thesis of one of this paper's authors, \citet{Schroeder2022PhD}, pays particular attention to the construction of KGs from such \emph{messy} enterprise data.
His proposed approaches are described in the following depending on the applied level.

\subsection{Corporate Level} \label{sec:KGC_corporate}
	Employees typically work on shared drives with enterprise-related and not personal documents.
	Prominent examples of such documents are spreadsheets.
	In the industrial sector, they are frequently used by knowledge workers to structure information since they are well-understood and provide an easy and fast way to enter data.
	However, precisely because spreadsheets do not predetermine how they should be filled, data is entered freely which causes the discussed messy data.
	For instance, it has been observed that people tend to insert data in spreadsheets in a ``sloppy'' way \cite{ConvertingAndAnnotatingQuantitativeDataTables} and that they frequently contain miscellaneous errors \cite{DBLP:journals/tois/BrownG87}.
	We would like to give a brief overview of possible challenges from our experiences \cite{SchroederJilekSchulze+2021inter}.
	If users enter multiple surface forms of the same entity, like ``Thomas Smith'' and ``Smith, T.'', the use of \emph{Natural Language Processing (NLP)} techniques such as named entity normalization \cite{NamedEntityNormalizationInUserGeneratedContent} becomes necessary.
	Comparable methods are required if, due to the reduction of typing efforts, acronyms and symbols are written by users to express entities and values.
	Similarly, mixed representations, prominently if dates are written, need to be unified, too.
	Since spreadsheets allow the usage of styled text, typographical emphasis, fonts and colors (background and foreground) may express certain information.
	Because of the free entering of text, it can happen that multiple entities are mentioned in a cell or multiple entity types are listed in a table.
	By mentioning (possibly ambiguous) identifiers, users express implicit references to entities or relationships between them.
	Such examples demonstrate what peculiarities have to be considered in case of messy spreadsheets.

	\subsubsection{AnnoSpreadKGC -- An Interactive Approach Using Annotated Spreadsheets}

\begin{figure}
		\centering
		\includegraphics[width=\linewidth]{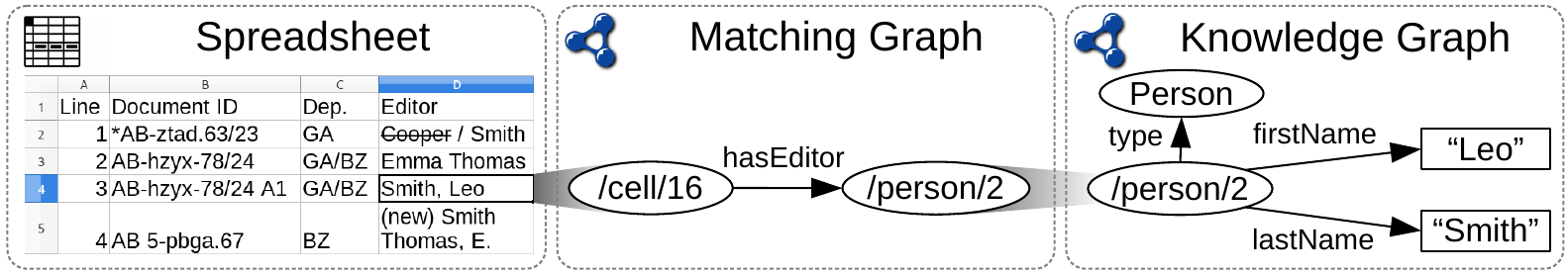}
		\caption{
			AnnoSpreadKGC annotation mechanism:
			spreadsheet (left) is annotated through a matching graph (middle) with extracted resources managed in a KG (right).
		}
		\label{fig:KGC_interactive_approach}
	\end{figure}

	To tackle these challenges, we proposed an interactive approach in \citet{SchroederJilekSchulze+2021inter} called \emph{Annotated Spreadsheet for Knowledge Graph Construction}, abbreviated \emph{AnnoSpreadKGC}.
	\Cref{fig:KGC_interactive_approach} illustrates its annotation mechanism: 
	a sheet (left) is explored and individual cells are selected by a knowledge engineer.
	Various methods can be applied on them performing knowledge extraction techniques and storing results in a KG (right). 
	Cells and KG resources are associated with a matching graph in the middle.
	In a final step, rows with their annotated cells are interpreted as entities forming a holistic interconnected graph.
	The approach's graphical user interface (GUI) provides several methods:
	Calculations of descriptive statistics gives an overview of categorical data in sheets and candidates for resources.
	Regular expressions are suitable to extract structured text information in cells.
	A date extraction method recognizes several formats and unifies them.
	The person index creator extracts mentioned persons and distinctly catalogs them.
	A membership discovery procedure reveals implicit relationships between cells.
	Using these methods $82$ times, we successfully turned five spreadsheets from an industrial scenario into a KG having $25,016$ triples about $2,719$ instances assigned to $15$ classes.
	However, because of manual configurations and operations in the GUI the approach involves considerable time and effort.
	This is also the reason why similar executions cannot be redone automatically, since there is no formalism that defines execution steps in form of mappings.
	
	\subsubsection{Utilizing the RDF Mapping Language for Spreadsheets}

	We therefore had the idea to enable the mapping of spreadsheets using the \emph{RDF Mapping Language}\footnote{
	  \url{https://rml.io/specs/rml/}}
	(RML) \cite{DBLP:conf/www/DimouSCVMW14}.
	Since no state-of-the-art RML mapper supports spreadsheets, we proposed and developed this feature in \citet{SchroederJilekDengel2021kgc}.
	By extending the Java-based engine \emph{RML Mapper}\footnote{\url{https://github.com/RMLio/rmlmapper-java}} with \emph{Apache POI}\footnote{\url{https://poi.apache.org}} for accessing Microsoft Excel files and a RML-compatible spreadsheet reference formulation, we are able to map spreadsheets to KGs using RML.
	The five sheets from our industrial scenario are mapped with RML using $ 25 $ logical sources, $ 26 $ triples maps, $ 126 $ predicate object maps and $ 51 $ function maps (utilizing FnO \cite{DBLP:conf/esws/MeesterDVM16}).
	However, only applying a cell-by-cell mapping turns out to be inadequate once NLP techniques for multiple entities with several surface forms or the discovery of implicit relationships involving global search and non-trivial matching needs to be performed.
	Additionally, the definition of correct RML mappings is time-consuming, especially if data needs to be inspected and discrepancies need to be recognized due to messy data.
	
	\subsubsection{Spread2RML -- Predicting RML Mappings}

\begin{figure}
		\subfloat[\label{fig:KGC_Spread2RML_E_img}A messy spreadsheet where two dates are in a cell and one is colored in red.]{\includegraphics[height=5cm]{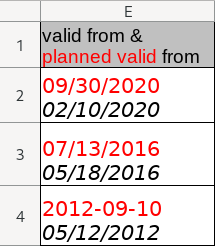}}
		\hfill
		\subfloat[\label{fig:KGC_Spread2RML_E}RML-based mapping to XSD date literals, but only red colored ones are considered with the \texttt{getEntitiesByColor} function.]{\includegraphics[height=5cm]{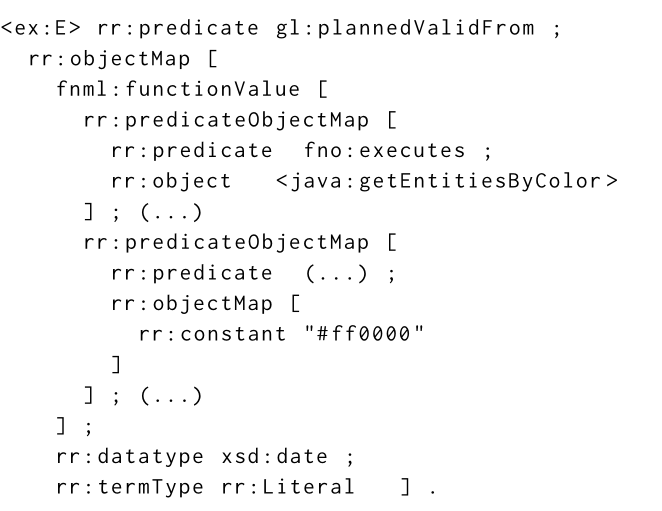}}
		\caption{
			Spread2RML analyzes a spreadsheet column (a) and suggests an RML mapping (b).
		}
		\label{fig:KGC_Spread2RML_examples}
	\end{figure}

	Therefore, we investigated the possibility of predicting RML mappings from spreadsheet data in \citet{SchroederJilekDengel2021s2r}.
	Our proposed approach \emph{Spread2RML} full-automatically suggests RML object maps by using an extensible set of predefined templates: currently fifteen are proposed in the paper.
	For each column, the most promising template is selected by calculating formally defined heuristics.
	To cope with messy data, eight FnO functions are defined to extract, parse and link cell values.
	\Cref{fig:KGC_Spread2RML_examples} illustrates a use case which involves style usage.
	An interactive demo with more examples is available online\footnote{\url{https://www.dfki.uni-kl.de/~mschroeder/demo/spread2rml/}}.
	In a data-driven experiment, we examined the performance of Spread2RML with three datasets: a synthetically generated one \cite{SchroederJilekDengel2021patterns}, a manually labeled dataset compiled from open government data\footnote{\url{https://www.data.gov/}} and five spreadsheets from our industrial scenario.
	\emph{Any23}\footnote{\url{https://any23.apache.org/}} (Anything To Triples) served as a baseline algorithm, because it converts CSV files to RDF automatically.
	Since datasets were labeled with expected KG statements, our evaluation is able to compare results and calculate precision, recall and f-measure values.
	For all datasets, Spread2RML outperforms the baseline and archives on average f-measure values between $0.27$ and $0.47$.
	Although, scores are comparably low, Spread2RML shows first promising results in the rather unexamined field of mapping recommendation for RML.
	
	\subsubsection{RDF Spreadsheet Editor}

\begin{figure}
		\centering
		\includegraphics[width=0.5\linewidth]{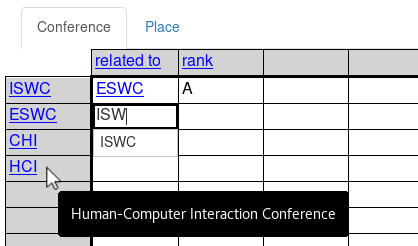}
		\caption{The RDF Spreadsheet Editor let users create resources (row header), properties (column header) and statements using intersecting cells.}
		\label{fig:HumL_RDFSpreadsheetEditor_img}
	\end{figure}

	Constructing a KGs from existing spreadsheets retrospectively can be a challenging task.
	For future sheets it could be beneficial if the KG is constructed \emph{while} data is entered by knowledge workers.
	This HumL concept is realized in our \emph{RDF Spreadsheet Editor} presented in \citet{SchroederJilekHees+2017,SchroederJilekHees+2018rdf}.
	While ontology editors need considerable technical knowledge and training to be of use, the spreadsheet metaphor is well-known and thus exploitable for transferring expertise.
	Using a fixed \textit{class per sheet}, \textit{entity per row} and \textit{property per column} schema as depicted in \Cref{fig:HumL_RDFSpreadsheetEditor_img}, users create resources and relationships mainly on an ABox level.
	Several features are proposed to increase its convenience, such as auto completion, copy \& paste, data type suggestion, shared modifications and an administration view.
	A user study with $17$ participants showed that users were able to create more statements in less time compared to using the ontology editor \emph{Protégé} \cite{DBLP:journals/aimatters/Musen15} or writing RDF in \emph{Turtle syntax} \cite{w3cRdfTurtle}.
	An additional user experience questionnaire \cite{DBLP:conf/usab/LaugwitzHS08} revealed an excellent attractiveness, efficiency and perspicuity score.
	Still, our editor can be further improved to handle multiple objects in a cell \cite{DBLP:conf/chi/BakkeKM11}, large graphs with searching, sorting and filtering capabilities and TBox support.
	
\subsection{Personal Level}
	Analyzing data assets or observing data entry, as shown for spreadsheets, give us the opportunity to grasp entities and their relationships in context of an enterprise.
	More personal information and work related concepts can be found on all devices employees work with.

    \subsubsection{PIM Crawler -- Crawling Personal Data}

	Data related to PIM is usually distributed in isolated applications on users' desktops.
	To gather all data to a single place for analysis purposes, we designed a dedicated tool called \emph{PIM Crawler}\footnote{\url{https://www.dfki.uni-kl.de/~mschroeder/demo/pim-semantifier/}}.
	Running on a user's desktop, it is able to collect information about files, calendars, emails and bookmarks in one relational database.
	Calendar data is acquired by reading \emph{iCalendar} \cite{rfc2445,rfc5545} files, while emails are collected through \emph{IMAP} \cite{rfc1064,rfc9051} or files in the Mailbox (mbox) \cite{rfc4155} format.
	Bookmarks can be extracted from web browser data stores, for example, from Firefox \cite{MozillaPlacesSqlite}.

    \subsubsection{Interactive Concept Mining on Personal Data}

\begin{figure}
		\centering
		\includegraphics[width=\textwidth]{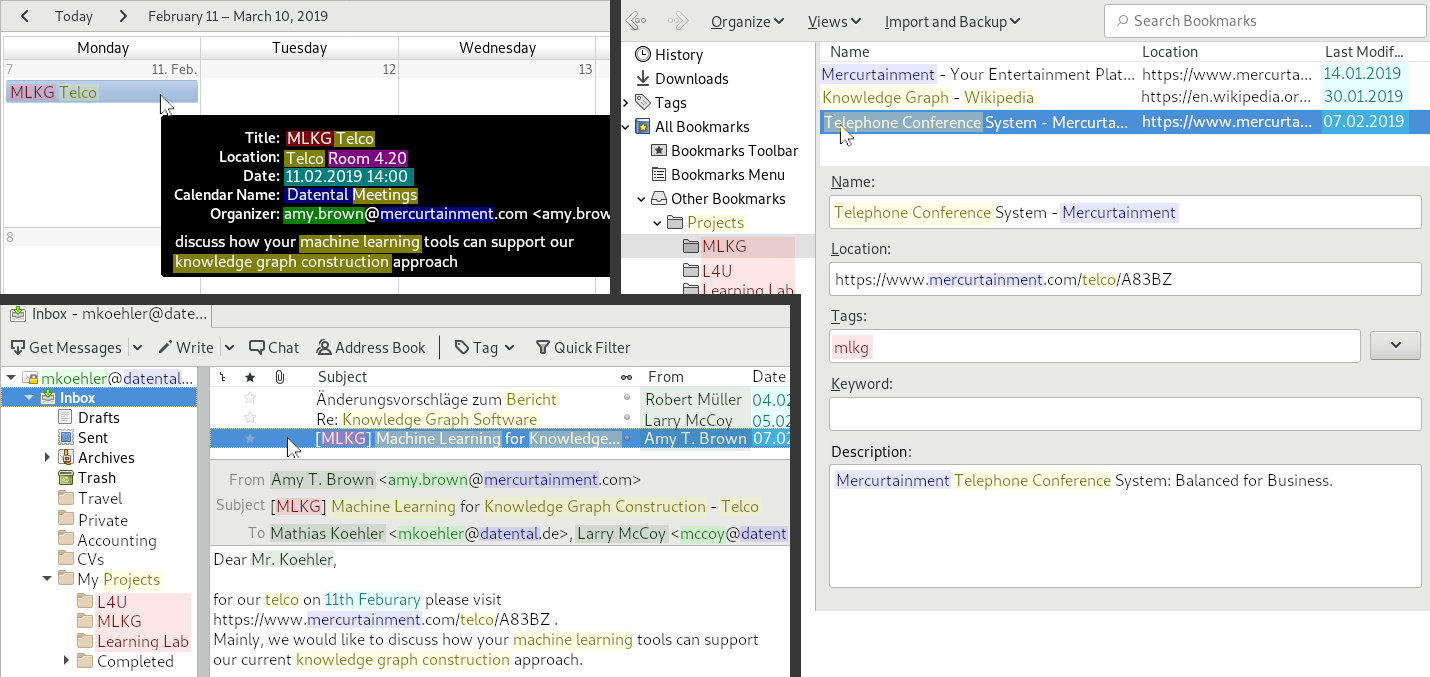}
		\caption{Promising concept candidates for PKGs which can be found in PIM structure of calendars, bookmarks and emails: persons (green), organizations (blue), projects (red), locations (purple), times (cyan), and general topics (yellow).}
		\label{fig:HumL_ConceptMining_Intro}
	\end{figure}

	Having access to personal information spheres of users allows us to find  higher level concepts in them such as persons, projects, and topics.
	To illustrate this, \Cref{fig:HumL_ConceptMining_Intro} provides an example of concepts we might find in calendar, bookmark and email data which could be promising candidates for a user's personal knowledge graph.
	Since users will always have a subjective view on the relevancy of such concepts \cite{Dengel2006,BergmanBeythMaromNachmias2008}, we proposed in \citet{SchroederJilekDengel2019} an interactive concept mining approach which exploits PIM structures and let users find and rate concept suggestions.
	Because relevant terms often come in various shapes, such as last names in mail addresses, frequent multi-word terms in email bodies or project names as upper-case acronyms, our GUI offers suitable ranking and filtering capabilities using metrics.
	These metrics score terms based on their occurrences in certain data fields as well as how deep or how distributed they occur in PIM structures.
	Yet, the approach does not learn from given user feedback and can overwhelm users with the high variety of settings.
	
	\subsubsection{Personal Knowledge Graphs from File Names}

\begin{figure}
		\centering
		\includegraphics[width=\textwidth]{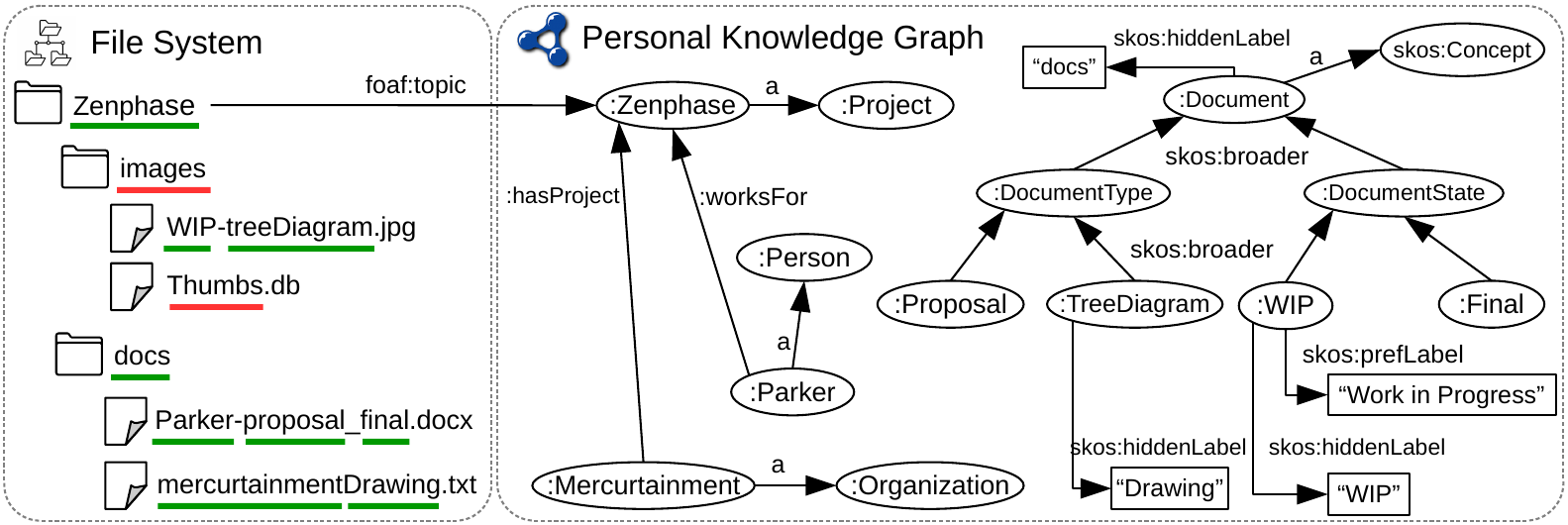}
		\caption{Relevant (green) and irrelevant (red) terms in file names (left) are used to construct a PKG (right) with taxonomic and non-taxonomic relations.}
		\label{fig:HumL_PKGFileNames_Example}
	\end{figure}

	Another promising source are files in hierarchical file systems, since users named them with their own vocabulary using task-related concepts, technical terms, made-up words and even puns \cite{carroll1982creative}.
	How file names can help in constructing personal KGs is illustrated in \Cref{fig:HumL_PKGFileNames_Example}:
	relevant terms in file names (underlined in green) lead to KG resources as well as taxonomic and non-taxonomic relations.
	An approach to construct such personal KGs from file names is presented in \citet{SchroederJilekDengel2022kgcf}.
	It supports the following construction tasks: domain terminology extraction, management of named individuals, taxonomy creation and non-taxonomic relation learning.
	The method makes use of rules and machine learning models to suggest new statements for the personal KG based on a user's feedback.
	A GUI is operated by a knowledge engineer, while a domain expert (data owner) is interviewed about the file system.
	We conducted four expert interviews (each for an hour) with four individual file systems to measure the performance of the approach.
	A questionnaire revealed that in fact file names reflect their language use and that on average, using a seven-point Likert scale \cite{Likert1932}, our system is able to construct personal KGs which meaningfully reflect their language, contain meaningful taxonomies and non-taxonomic graphs.
	The prediction performance of our models reached mixed accuracy values depending on the task which indicates much room for improvements.
	Yet, an effort estimation for our GUI showed that already two inputs are enough to express a true or false statement.
	
	\subsubsection{Deep Linking Desktop Resources}

	Having personal KGs with personal concepts and their relationships help AI systems to understand the employees' work-related domains.
	However, not all native documents on their desktops are associated with these concepts as intended by the Semantic Desktop vision \cite{SauermannBernardiDengel2005}.
	To bridge this gap, we proposed a \emph{deep linking approach} in \citet{SchroederJilekDengel2018}.
	Similar to \emph{fragment identifiers} \cite[Sec.~3.5]{rfc2396}, our method allows users to refer to certain locations within their files using deep links, which could be, for instance, a shape in a slide of a presentation.
	By enabling the association of resolvable deep links to files with concepts from KGs, we form richer PKGs with references to native data.
	
\subsubsection{Contextifier -- Retrospective Context  Mining on Personal Data}
Based on the aforementioned PIM Crawler and Concept Miner, the \emph{Contextifier} presented in \citet{HeimJilekMaus+2022} tries to identify user contexts spread across the different data systems.
For example, a file could have been the attachment of an email, things stored in a certain folder all belong to a specific meeting mentioned in the calendar, a bookmark in the web browser originated from an email the user has received, etc.
While many related approaches (see \Cref{sec:RelatedWork}) do context/task mining on-the-fly on a running system, some, like \citet{CostacheGaugazIoannou2010}, also take past data into account making them hybrid approaches.
To the authors' best knowledge, the Contextifier is the only purely retrospective context mining approach.
A reason for this could be the difficulty of the problem.
For example, from possibly millions of interactions with a file only a few timestamps remain: the creation, the last modification and the last access date (some of them may even be missing depending on the file system).
Since usually most of the interaction data is lost, such an approach is very speculative.
We therefore again follow a HumL approach.
Contextifier tries to identify relations between information items (see examples above) assigning scores in several dimensions to each pair, e.g. text similarity, similar found concepts in the items, closeness in a hierarchy like the file system, etc.
The core problem can be modeled as a multi-graph clustering problem having the information items as nodes and the different comparison results as (multiple) weighted edges between them.
Contextifier comes up with an initial suggestion that can be refined by the user in various feedback and re-calculation iterations.
Results of an early evaluation of the system involving 14 participants were promising \cite{HeimJilekMaus+2022}.\\

An initial bootstrapping phase (as described above) prevents (or at least reduces) the aforementioned \emph{cold start problem} an assistance system would suffer from.
The obtained KG as well as live connections to existing systems are a good data foundation for a \emph{Personal Knowledge Assistant (PKA)} to operate on.
This next section presents \emph{Self-organizing Context Spaces (cSpaces)} as an example for such a PKA.
\section{Self-organizing Personal Knowledge Assistants}
\label{sec:PKA}
Self-organizing Context Spaces, or \emph{cSpaces} for short, were mainly developed as part of a PhD thesis of one of the authors (\citet{Jilek2023PhD}) from 2018 to 2022 -- its development is ongoing for further research.
It is a variant of a self-organizing PKA and actually not (yet) a single application containing all features mentioned in the following.
Instead, features were developed step by step over several years and also re-combined or tailored towards certain research questions.
Especially before work on cSpaces had started, experiments were conducted with the already productively used \emph{CoMem} system.
Over the years, CoMem, whose development started in 2011, further matured coming closer to an industry-ready product (e.g. see Section \ref{sec:Industry}).
This also made it necessary to ``outsource'' certain experiments and research questions to more lightweight, experimentation-friendly environments that are typically also more ``fragile'', i.e. less stable and maintained.
cSpaces can thus be seen as a reduced, experimentation-tailored version of CoMem paving the way for potential future CoMem features.

In the remainder of the section, we first present the basic interaction cycle with systems like cSpaces or CoMem, followed by an overview of different user interfaces.
Next, we share our insights on how working with and in \emph{Context Spaces}, i.e. context an explicit interaction element, was perceived by test users so far and which benefits we could already identify or even quantify, respectively.
This is followed by an overview of support measures offered by CoMem and cSpaces.
Last, searching and trust in such self-organizing and especially \emph{Forgetting-enabled Information Systems (FIS)} are addressed.

\subsection{Interaction Cycle}
The interaction with cSpaces is basically a cycle of six steps \cite{Jilek2023PhD}:

\begin{enumerate}
\item \textbf{Evidence collection}:
First, there is evidence collection, i.e. the system tracking user behavior like reading/writing files, clicking websites, etc.
In Semantic Desktops, there are basically two types of applications: native ones that were especially created to work with the Semantic Desktop or plug-ins to enhance traditional applications.
We follow both approaches.
However, for the latter, typical plug-ins also contained a front-end written in the particular app, which often resulted in increased development effort.
We reduced this effort by creating \emph{tiny headless plug-ins} that simply \emph{send out} in-app events to the Semantic Desktop -- we therefore called them \emph{plug-outs} \cite{JilekSchroederSchwarz+2018}.
More details on such plug-ins as well as Semantic Desktop architecture and re-engineering using plug-outs can be found in \citet{Jilek2023PhD}.\\

\item \textbf{Information extraction}:
Next, the collected user activity evidence snippets are analyzed using, for example, information extraction methods.
We mainly perform a variant of ontology-based Named Entity Recognition (NER), in particular identifying mentioned entities of the KG in the snippets.
Items like files, email, websites, etc. are then automatically annotated with a \emph{hasSuggestedTopic} relation, e.g. <DFKI website> \emph{hasSuggestedTopic} <machine learning>.
Since our system is a real-time assistance system, this analysis needs to be performed as fast as possible:
According to \citet{Miller1968} and \citet{CardRobertsonMackinlay1991}, as cited in \citet{Nielsen1993}, 100 ms is ``about the limit for having the user feel that the system is reacting instantaneously'' and 1000 ms is ``about the limit for the user's flow of thought to stay uninterrupted''.
Since our system also needs some time for calculating and performing the support measures that follow the analysis, the actual NER process should only take a small two-digit number of milliseconds on typical end user hardware.
Additionally, a certain robustness regarding lexical variation of entity names is required, so the system does not ``lose track'' of users' activities by not detecting mentioned entities.
This is especially relevant for highly inflectional languages like German, Spanish, Latin, Hebrew, Hindi, Slavic languages, etc.
In \citet{JilekSchroederNovik+2019}, we presented a specialized NER approach combining (near-)real-time requirements with inflection tolerance.\\

\item \textbf{Context elicitation}:
The third step is context elicitation.
Given the evidence snippet(s) and the analysis result, the system tries to identify reasonable context candidates.
Since users are directly interacting with the Semantic Desktop like explicitly selecting a context to work in or dragging items into a context space, the task can also be stated as predicting whether the user is still in the same context or whether they switched to another one (and if that is the case, to which one).
Several context elicitation methods were presented in \citet{Jilek2023PhD}.\\

\item \textbf{Information value assessment}:
In the fourth step, Memory Buoyancy and Preservation Value scores are calculated (see later parts of this section).\\

\item \textbf{Self-(re)organization measures}:
Based on these scores, support measures may be triggered like temporal hiding, reorganization, condensation, etc.
Details will follow in the remainder of the section.\\

\item \textbf{User interface updates}:
Steps 5 and 6 are the actual support measures, whereas step 5 can be seen as the back-end part and step 6 addresses the front-end, i.e. updating the GUI appropriately.
\end{enumerate}

\begin{figure}
  \centering
  \includegraphics[width=1\columnwidth]{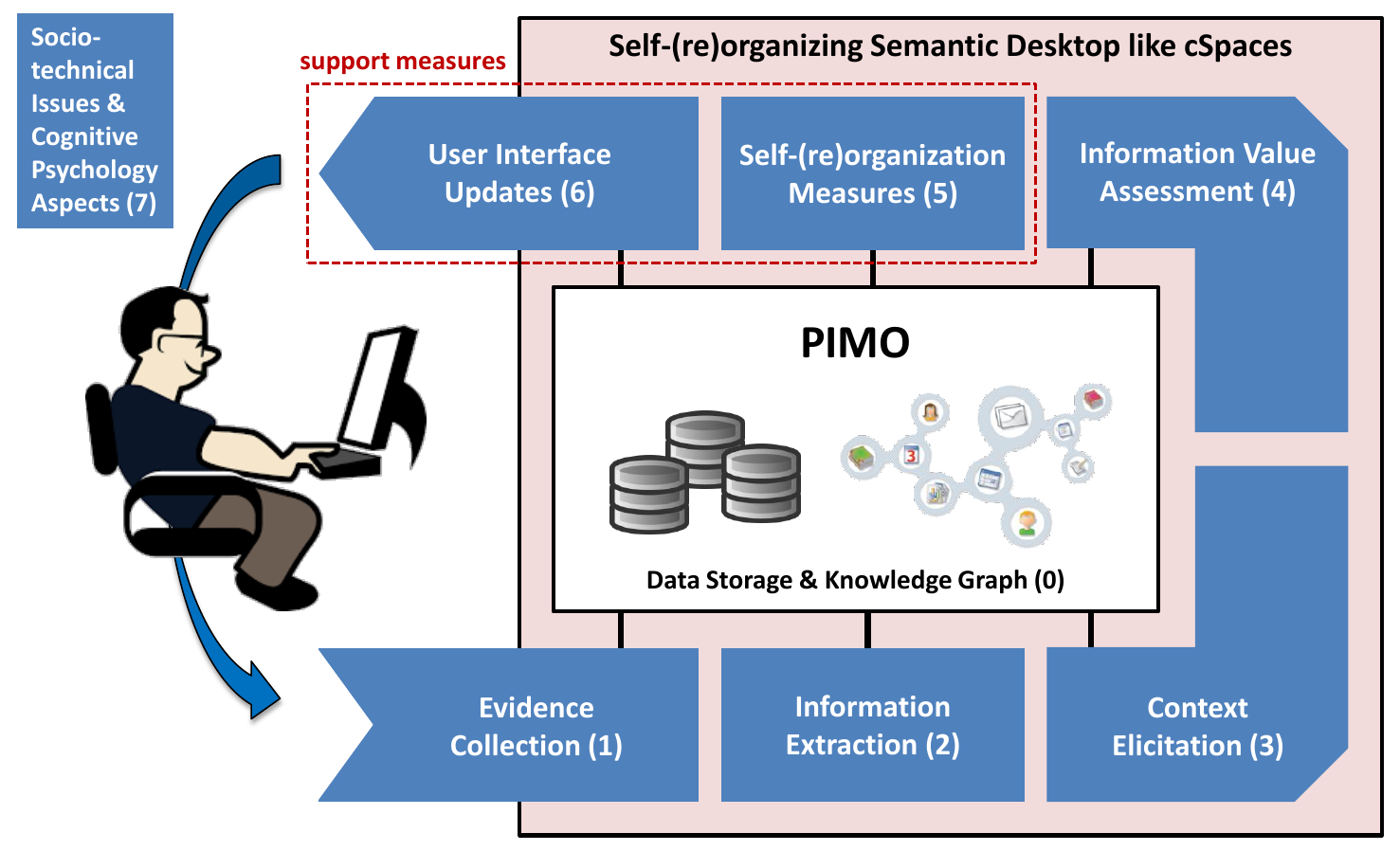}
  \caption{cSpaces interaction cycle}
  \label{fig:interactionCycle}
\end{figure}

After step 6, the interaction goes into its next iteration back at evidence collection.
The interaction cycle is depicted in Figure \ref{fig:interactionCycle}.
There are two additional aspects to be mentioned:

\begin{enumerate}
\setcounter{enumi}{-1}
\item \textbf{Data storage and knowledge graph (PIMO)}:
First, the data foundation (PIMO as a personal KG, databases, etc.) mentioned in the last section are drawn as step/aspect 0 in the center of the figure.
The aspect not only comprises data storage and connection to other systems but also context modeling.
cSpaces extends the context model by Schwarz \cite{Schwarz2005,Schwarz2006}, which itself is an extension of the one by Maus \cite{Maus2001}.
Both researchers already introduced \emph{organizational} (e.g. role of the user), \emph{historical} (previous tasks), \emph{causal} (tasks, goals), \emph{informational} (recent documents and topics), \emph{operational} (active applications and tools), \emph{behavioral} (user actions), \emph{environmental} (location, hardware) and \emph{attentional aspects} (text scope, read/skim text) into the model, which Jilek extended by additionally introducing \emph{context hierarchy} (sub-/super-contexts), \emph{forgetting} (forgotten parts of a context) and \emph{focal aspects} (user's last focus, i.e. items interacted with, in a context) \cite{JilekRungeNiederee+2019,Jilek2023PhD}.
Storing and querying all this data imposes the challenge of combining structured and indexed search in near-real-time scenarios of highly evolving data (every click of the user may change the KG, the contexts, etc.), which is a large topic of its own.
In \citet{Jilek2023PhD}, requirements for such a data store were summarized and an early prototype was developed and presented.\\

\setcounter{enumi}{6}
\item \textbf{Socio-technical issues and cognitive psychology aspects}:
Together with research partners of cognitive psychology and ergonomics, we also investigate socio-technical issues of such an assistance system.
\end{enumerate}

Having introduced the basic interaction cycle with the system, the next section presents some of its user interfaces.

\subsection{User Interfaces}  \label{sec:userInterfaces}
\begin{figure} 
  \centering
  \includegraphics[width=1\columnwidth]{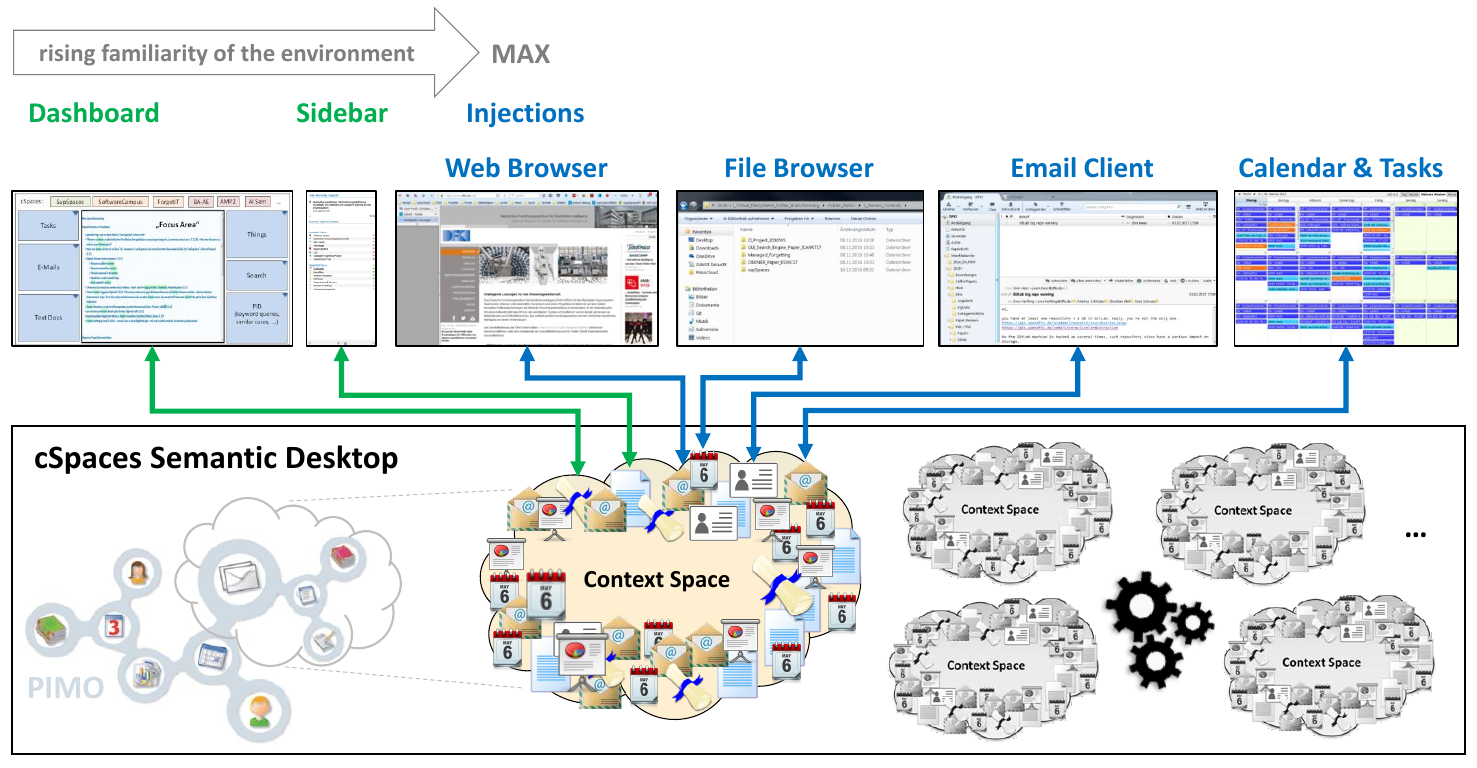}
  \caption{cSpaces user interfaces}
  \label{fig:userInterfaces}
\end{figure}

So far, three different user interfaces have been realized for cSpaces: a dashboard, a sidebar and transparent injections into existing systems like file or web browsers.
As shown in the upper left of Figure \ref{fig:userInterfaces}, each of these interfaces offers a bit more familiarity to the user.

\textbf{Dashboard}.
With the potentially screen-filling size of a \emph{dashboard}, there is lots of space to visualize information and offer interaction possibilities for search, tagging, etc.
There is, however, the drawback of having to leave the current application and switch to another app (the dashboard) typically taking up the screen's whole space obscuring the user's view on all or most other applications.

\textbf{Sidebar}.
A \emph{sidebar} mitigates this problem by occupying far less space but, on the downside, all widgets need to be more compact to fit into the sidebar, which typically means dropping visualization or input elements that would have been shown in the dashboard.

A dynamic solution called ``\emph{sashboard}'' was proposed in \citet{Jilek2023PhD}: users basically work with a sidebar and whenever they need more space, they drag the inner side of the sidebar towards the center of the screen as if closing a (horizontal) sash window.
The sidebar then re-adjusts itself to a dashboard that is filled and pre-configured with the current context space's content.
For example, a small quick search widget resizes into a full-fledged faceted search interface.

\textbf{Transparent Injections}.
The highest familiarity with the environment is achieved using \emph{transparent injections} of cSpaces into existing systems like the file, mail or web browser, calendar etc.
As presented in \citet{JilekSchroederSchwarz+2018}, cSpaces uses standard protocols like \emph{SMB} \cite{HeizerLeachPerry1996,MicrosoftSMB2} (files), \emph{IMAP} \cite{rfc1064,rfc9051} (mails) or \emph{WebDAV} \cite{rfc2518,rfc4918} and its extensions \emph{CalDAV} \cite{rfc4791} and \emph{CardDAV} \cite{rfc6352} (calendar, address book) or similar quasi-standards like \emph{WebExtensions} \cite{MozillaWebExtensions} (web and mail browser),
supported by all major operating systems to inject contexts on a very low level without touching any operating system code.
Especially the file system is interesting in this regard since it is available in all applications and thus makes cSpaces available in all of them.

\begin{figure}
  \centering
  \includegraphics[width=1\columnwidth]{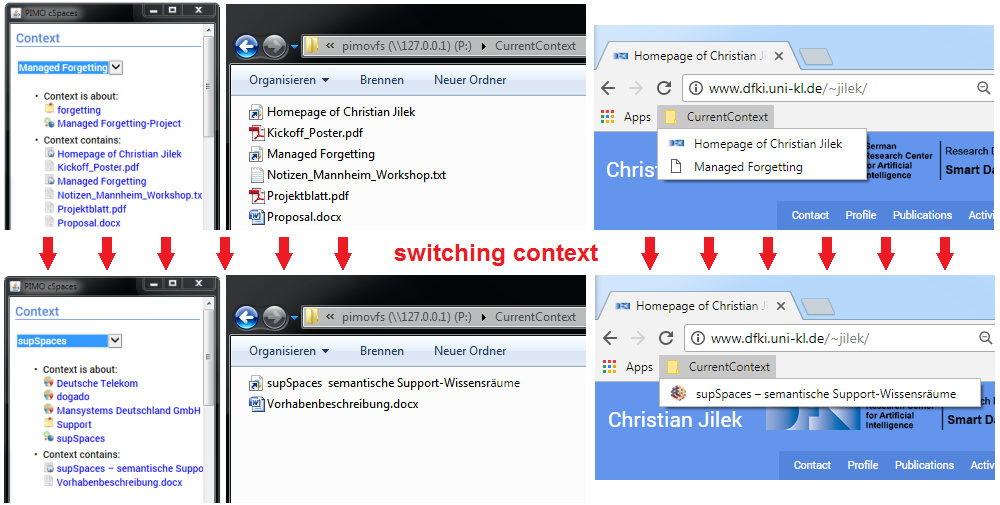}
  \caption{Context switching with cSpaces}
  \label{fig:contextSwitching}
\end{figure}

Higher-level actions like tagging items or writing comments could become cumbersome using only these injections.
Therefore, a sidebar or dashboard is used as a complement.
Figure \ref{fig:contextSwitching} illustrates how working with cSpaces then looks like.
In the upper half, we see the \emph{Managed Forgetting} project being the context currently selected in the sidebar (left).
The context is associated with two concepts (``is about''), and, in this example, it contains six items, four files and two bookmarks.
We see that this context space (``current context'') has been injected into the file system (center) and the web browser (right), whereas the web browser only shows the two bookmarks.
In its lower half, the figure depicts what happens if a user switches to another context: the content of the sidebar is replaced with the new context's information and both injections are updated accordingly, i.e. files and bookmarks are replaced.
Since the standard protocols are used like a virtual layer, no actual data on the harddisk is moved making this operation executable in typically very few milliseconds.

\subsection{Working with and in Context Spaces}
One major idea of cSpaces is to enable users to actually work \emph{with} (i.e. a ``tangible object'' similar to a folder) and \emph{in} (immersion) context spaces.
This especially allows for unified browsing of contexts that are close(r) to a person's mental model and easy ways of tagging/associating items with these contexts.
``Modeling'' (in the sense of users making more of their mental model explicit for the machine) should be easy and subtly possible for the user, e.g. dragging a file to a context makes the system automatically create an \emph{isContainedIn} relation.
If the context represents a calendar event, e.g. a business meeting, then the meeting can be associated with entities found in that file, etc.
cSpaces addresses all three \emph{pressing requirements for future PIM systems} stated by \citet{Warren2013}: combat information overload, ease context switching and support information integration across a variety of applications.

\begin{figure}
  \centering
  \includegraphics[width=1\columnwidth]{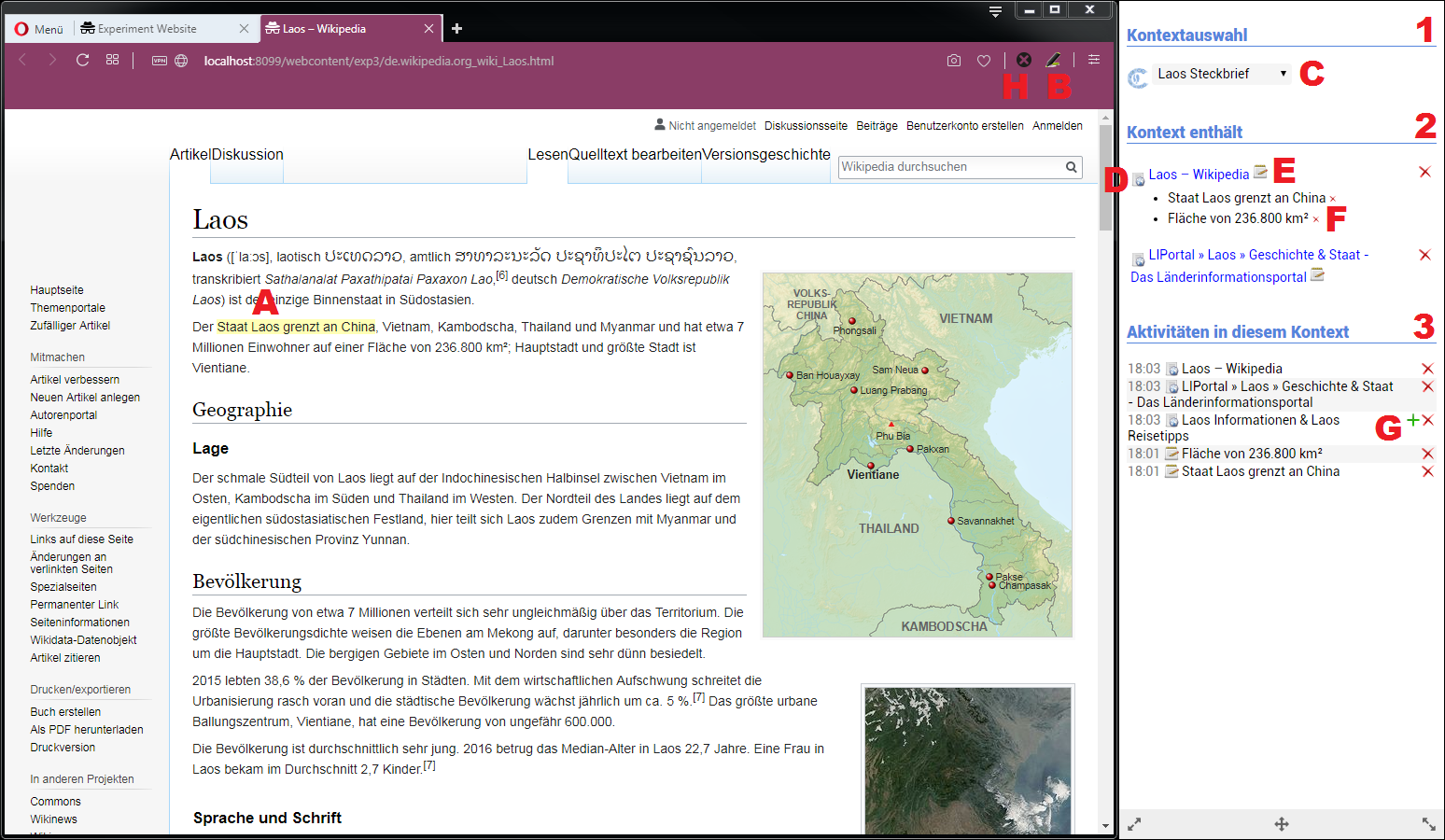}
  \caption{cSpaces prototype for several user studies}
  \label{fig:prototype}
\end{figure}

We have already conducted multiple studies to gather user feedback on the system and to quantify potential benefits.
Figure \ref{fig:prototype} shows a typical experimentation scenario with cSpaces: users interact with the sidebar (right) next to their usual applications, in this case a web browser (left).
In this example, the sidebar consisted of three parts: the context selection (1), a ``context contains'' section filled with the items of the selected context (2) and an activity history showing the (last) actions performed in the context in reverse-chronological order (3).
The screenshot particularly shows a web research scenario, in which participants investigated a certain topic, e.g. information about a country or famous person.
They could add items (webpages) to the currently selected context (C) by clicking on the plus symbol next to the webpage's entry in the activity history (G).
The webpage then appeared in the ``context contains'' area (D).
They could also write notes associated with the webpage (E) or use a highlighter in the web browser (A and B), whose selected text passages were automatically converted into notes (F).
To ensure that all browser tabs (except for the experiment instructions) were closed when participants were, for example, asked to switch contexts, a special tab closing app was developed (H).
This prototype or slightly varying ones were used in several studies presented in the following.

\subsubsection{Cognitive Offloading Effects}
In a first study, participants were asked to do web research on a certain topic.
The cSpaces assistant in a variant as described above captured their respective context, they could take notes, etc.
In the middle of their task (after a few minutes), they were interrupted to switch to another completely unrelated context, the solving of modular arithmetic problems by \citet{Bogomolny1996} (referring to \citet{Gauss1801}).
These equations are known to ``place heavy demands on working memory'' as stated by \citet{BeilockCarr2005} referring to \citet{Ashcraft1992,AshcraftKirk2001}.
Before switching to the math context, participants were informed whether their previous context (web research) would be kept or deleted.
In case of the former, they could rely on the system keeping their context and bringing it back up when needed later.
Participants whose context was kept were actually able to solve more equations than the deletion group (10.3 vs. 9.4, $n=48$, $p<0.01$).
Although it was only a weak effect, the difference in means was significant.
Participants whose context was kept were able to ``cognitively offload'' \cite{RiskoGilbert2016} their so far collected facts in the research task to cSpaces freeing working memory capacity, which then was additionally available for the math task.
Our colleagues of cognitive psychology had already shown the effect in a scenario of encoding and recall of word material \cite{TempelFrings2016}.
With this experiment, we were able to transfer these findings to a more complex scenario like knowledge work.
For details please see \citet{GauselmannRungeJilek+2022}.

In a follow-up experiment \cite{GeisslerGauselmannJilek+2023}, we could show the same effect more directly using \emph{functional Near-Infrared Spectroscopy (fNIRS)}, ``a neuroimaging technology for mapping the functioning human cortex'' \cite{FerrariQuaresima2012}.

\subsubsection{Effects on Task Resumption/Switching}
\begin{figure}
  \centering
  \includegraphics[width=1\columnwidth]{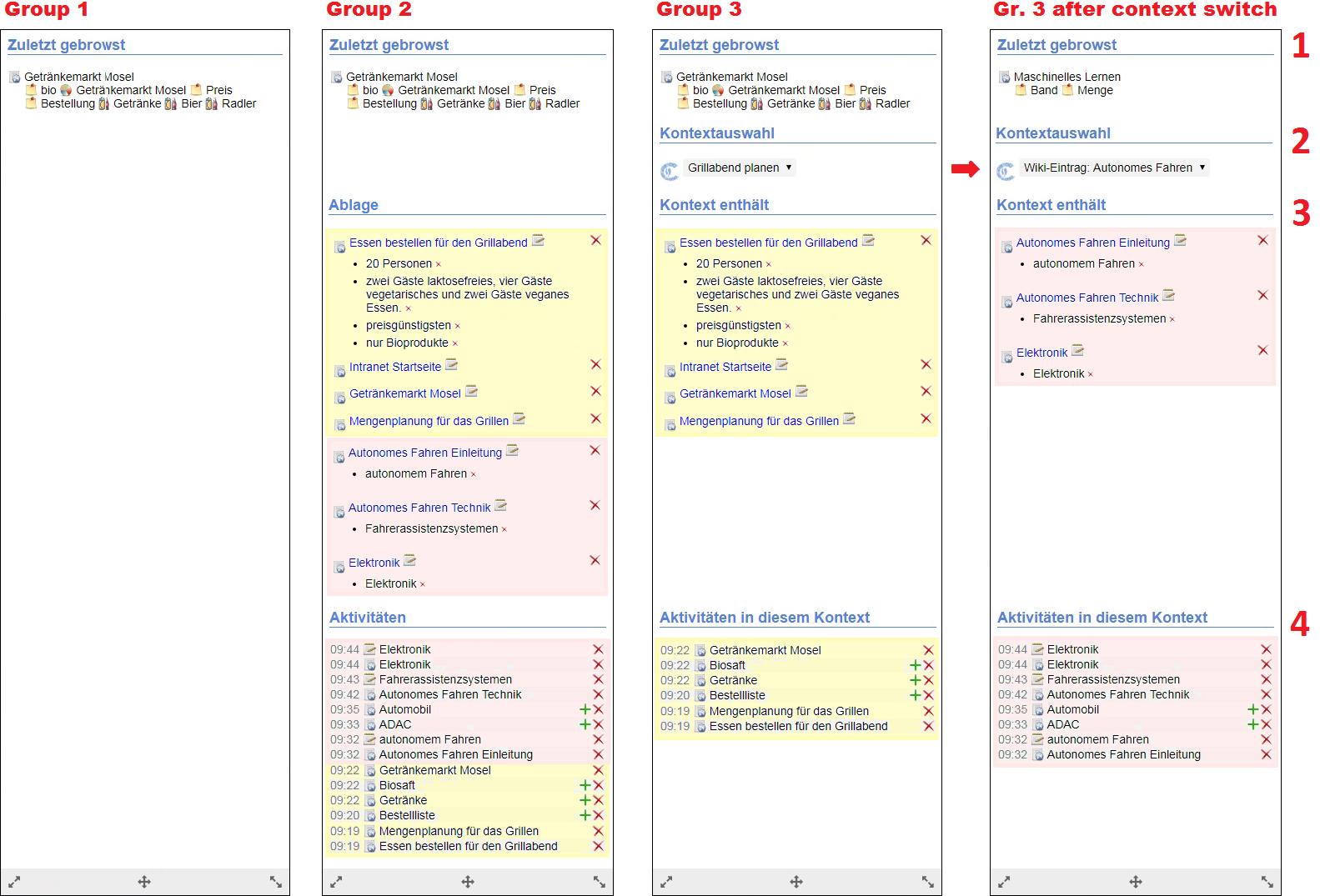}
  \caption{Task switching with three different variants of the cSpaces sidebar}
  \label{fig:taskSwitching}
\end{figure}

In a second study, we investigated effects on context/task switching with the goal of quantifying participants' so-called \emph{Task Resumption Lag (TRL)} \cite{AltmannTrafton2004}, which we understand as the time a person needs to re-adjust their mind to the former context in order to continue where they left off.
Participants were asked to plan a barbecue evening that made them do research like checking emails by guests, in which they stated what they would like to eat or bring for the party, vendor websites with prices, etc.
Step by step, they had to fill an order list with amounts of food and beverages to be ordered.
From time to time (in each case after a few minutes), they were interrupted to work on an unrelated task (working on a wiki entry about autonomous driving).
Each participant was interrupted and a few minutes later resumed the main task three times.
As shown in Figure \ref{fig:taskSwitching}, the experiment had three groups:
Group G1, a control group, had a non-interactive sidebar: it only showed a concise content analysis of the currently browsed email or website (1), basically the most prominent entities found in the text.
Group G2 additionally had the possibility to add items to a single context-free folder as well as adding notes to them (3).
Their activities were captured in an activity history list (4).
Group G3 also had these possibilities but with the additional feature of being able to switch contexts (2).
As a consequence, items, notes and activities stayed separated by contexts and were not mixing up.
Figure \ref{fig:taskSwitching} shows how content of the two contexts -- from the experimenter's view: main task and distraction task -- colored in yellow and red, respectively, mix up in the sidebar of G2, whereas they stay separated for G3.
After an initial tutorial to get to know the system's different features, each participant worked in total for 20 minutes on the main task and 12 minutes on the distraction task.
The experiment contained 40 different webpages -- emails were shown as part of an online mail client, so the whole experiment could be conducted using just a web browser.
With the experiment having 51 participants, we captured 153 observed context switches as well as one hour of video material and $1,000$ logged activities on average per participant.
Participants were not asked to explicitly state when they felt their switch back was complete (mainly due to the experiment initially having a slightly different focus; however, the rich amount of obtained data could also be used for the TRL analysis in a secondary study).
The TRL was measured by analyzing the captured evidences.
It was assumed that a person was back in their former context 1) when getting back to pages actually containing content needed to fill the remaining lines of the order list or 2) adding new items or notes to the current context.
To detect 1), all 40 web pages used in the study were classified as being either offerings and prices, amounts of food or beverages brought or requested by guests, navigation pages, spam without any relevant information or the order list to enter results.
Next, this classification was applied to the activity logs in order to calculate the TRL for each of the 153 context switches.
Last, these results were compared to the captured video material and corrected where necessary.
For instance, this was the case when users navigated through several pages very quickly, staying on each page for less than a second.
Even if a page classified as content was among those, it was assumed that the detention time was too short to actually conceive anything written on that page.
Thus, the next content page visit or sidebar addition in the activity log was used to calculate the TRL.

The average TRLs for each group were as follows: 18.3 seconds for G1, 11.0 seconds for G2 and 10.5 seconds for G3.
cSpaces-supported groups were thus about 40\% faster in resuming their former context than the control group.
The small difference between G2 and G3 (not significant) can be explained by only having two different contexts, of which one even was a distraction task and discovered as such by some participants:
we observed rare cases of G2 participants taking the time to tidy up their sidebar by removing items and notes after returning from the distraction to the main task.
It is worth noting that in post-experiment interviews, 78\% of G2 participants stated (or at least insinuated) the need for a context-sensitive sidebar not mixing things up.
More contexts would have presumably been necessary to see a larger TRL difference between G2 and G3.
However, the differences of G2 and G3, respectively, to G1 were significant ($p<0.001$) and can be classified as a strong effect.
Further details on the experiment can be found in \citet{JilekGauselmannChwalek+2021} and \citet{Jilek2023PhD}.

\subsubsection{Further Insights of a Multi-month User Study}
As a complement to the previously mentioned short-term studies, we set-up a multi-month user study, in which a smaller group of participants could use a more advanced prototype for up to five months (some users joined later in the study).
The group of participants consisted of seven fellow researchers and students.
They were asked to integrate the application into their daily work, which let to some of them using the app nearly every work day and others only a few days per month.
In total, 249 days of use could be captured, consisting of $46,552$ individual user events, 173 context spaces, 165 browsed files and $8,393$ browsed webpages.
The study ended with a structured interview of 45 questions on 15 aspects of cSpaces.
For example, all participants clearly agreed that
\begin{itemize}
\item the ideas behind cSpaces were easy to understand,
\item learning how to use the app was easy and its usage could easily be remembered,
\item taking notes on items in a context was exactly where they would like to have such notes,
\item switching between different user interfaces like dashboards and sidebars would help in being more efficient and
\item that they had not seen something like cSpaces in existing software applications.
\end{itemize}
Besides rating and describing their experience with the current prototype, they were in some cases also asked to extrapolate their gained experience to envisioned or early features they knew about (e.g. by seeing them in demos) but which were not part of the current prototype, yet.
Such items were for example:
\begin{itemize}
\item whether items were right where participants would like to have them using context spaces,
\item whether such context spaces were more aligned with their mental model than with traditional systems,
\item whether cSpaces would allow for faster context switches,
\item whether cSpaces helped in keeping their information sphere more tidied up or
\item cSpaces showing only the current context while temporarily hiding others reduced cognitive load while working.
\end{itemize}
More questionnaire items will be mentioned in the subsections about the different support measures.
Being a research prototype, the cSpaces application still had usability issues that were less problematic in short-term experiments.
However, in a more productive medium/long-term use, some ``quality of life`` improvements would have been helpful, e.g. clustering of the activity history, introducing better scrolling for certain widgets, etc.
Note that these were not conceptual problems: users could give separate ratings for the current and the envisioned prototype and some problems were clearly attributed to the current but not the envisioned one.
In general, participants were asked to give their honest opinions and ratings, and they actually pointed out the problems above.
Thus, a courtesy bias \cite{LeonLundgrenHuapaya+2007}, i.e. known or friendly persons giving more positive answers than they would have with an unknown experimenter or the prototype of an unknown person, can probably be ruled out for this study.

In summary, one may conclude that the current prototype was already perceived as helpful (in total, ratings were slightly on the positive side of the spectrum), whereas even more potential was seen in the envisioned or early features (ratings very much on the spectrum's positive side).
For a detailed break-down of all results please see \citet{Jilek2023PhD}.

\subsection{User Support Measures}
This section gives more insights into user support measures offered by cSpaces and/or CoMem.
They are an implementation of Managed Forgetting comprising ideas like condensation, temporal hiding or reorganization (see Section \ref{sec:Background} for the complete definition).

\subsubsection{Condensation and Summarization}
\textbf{PIMO Diary}.
An example of this category is \emph{PIMO Diary}, a diary generator proposed in \citet{JilekMausSchwarz+2015}.
It takes a person's personal KG (PIMO) and a given timespan as inputs and tries to cluster items and events in a way that makes the diary interesting to read/browse.
There should be, for example, a certain amount of diversity within the diary, i.e. events belonging to the same project or topic should rather become one cluster instead of several ones about more or less the same.
The app generates headlines for each entry, shows associated media like photos, the most important associated concepts of the KG as well as a word cloud of the most important terms.
Users can zoom-in and -out or shift the focus towards certain topics.
PIMO Diary is basically a clustering approach using text and concept embeddings.

A future idea for PIMO Diary could be to remove (i.e. archive or delete) parts of the KG and instead have a diary entry serve as a condensed \emph{memory landmark} \cite{RingelCutrellDumais+2003,HorvitzDumaisKoch2004} summarizing the respective sub-graph.
Only if needed again (i.e. Memory Buoyancy rising), the condensed version would be unfolded back to the full graph (if details were archived).

\textbf{Flat Context Views}.
Another condensation idea are \emph{flat context views} \cite{Jilek2023PhD} for context hierarchies similar to flat package views known from software development environments.
The idea is to hide certain sub-contexts and several/most of their items presenting only the most relevant parts of the whole context hierarchy at a glance.

\subsubsection{Temporal Reorganization, Fading Out and Resurfacing}
\begin{figure}
  \centering
  \includegraphics[width=1\columnwidth]{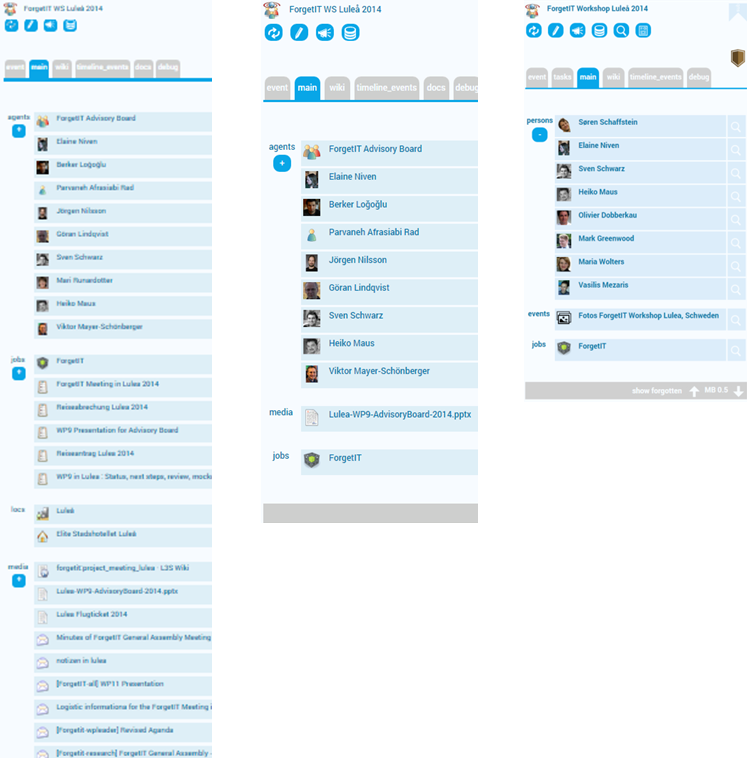}
  \caption{Fading out using the example of a calendar entry viewed during the event (left), eight months later (middle) and two years later (right)}.
  \label{fig:fadingOut}
\end{figure}

\textbf{Fading out and Resurfacing in CoMem}.
A feature directly driven by Memory Buoyancy (MB) is \emph{Fading out and Resurfacing}.
MB tries to assign a score to each information item that represents the current value for the user.
It was created according to findings of human memory and cognition and its design principles in CoMem were presented in \citet{MausJilekSchwarz2018}.
A first evaluation was conducted in \citet{TranSchwarzNiederee+2016}.
Later, \citet{JilekChwalekSchwarz+2019} presented an advanced version to better cope with different contexts and sudden switches.
A selection of the aforementioned design principles is as follows (for a full list please see \citet{MausJilekSchwarz2018} and \citet{JilekChwalekSchwarz+2019}):
\begin{itemize}
\item MB is updated every time an item or thing in the KG is stimulated; stimulation spreads into the sub-graph using \emph{spreading activation} \cite{Crestani1997}; the stimulation strength depends on interaction type, type of the thing and connections in the KG.
\item There are measures to cope with activity bursts and erratic accesses: single access should not lead to high MB, multiple accesses in short time are treated reluctantly and multiple accesses over a day saturate against a high MB.
\item MB drops for things that are not stimulated: first a steep decline, then a long-tail of slow decline.
\item Upcoming events should stimulate connected things; finished things (events, tasks) shall decrease faster unless other indicators speak against this; times with generally low user interaction should not lead to massive decay in MB.
\end{itemize}
Figure \ref{fig:fadingOut} shows MB ``in action'' using the example of a calendar entry.
The figure especially shows the fading-out part: all details of the event were visible during the event (left).
Steadily over time, details start to fade out: the center screenshot shows the same event after eight months (emails and tasks now hidden by default; presentation slides remain visible) and the right one after two years (especially the event's photo collection remains visible, presumably since colleagues were reminiscing every now and then).

\textbf{Context Overview}.
Other features with regard to temporal hiding or reorganization are a \emph{context overview} that in an MB-like way brings up only the most important contexts first.
Contexts that drop in relevance are hidden from the default view, which a user possibly views each morning when entering the office and starting their computer.

\textbf{Last Focus}.
Saving the \emph{last focus} of a context is another idea, i.e. saving open/active applications, the last persons a user was in contact with in a certain context, etc.
Since MB is a multi-dimensional value and not a simple \emph{least recently used} timestamp, such focal items may not automatically be the ones with high MB.

\begin{figure}
  \centering
  \includegraphics[width=0.4\columnwidth]{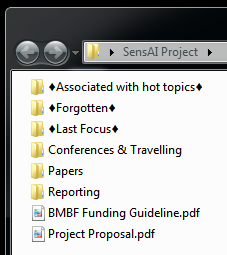}
  \caption{Intelligent folder injections}
  \label{fig:IFIs}
\end{figure}

\begin{figure}
  \centering
  \subfloat{\includegraphics[width=1\textwidth]{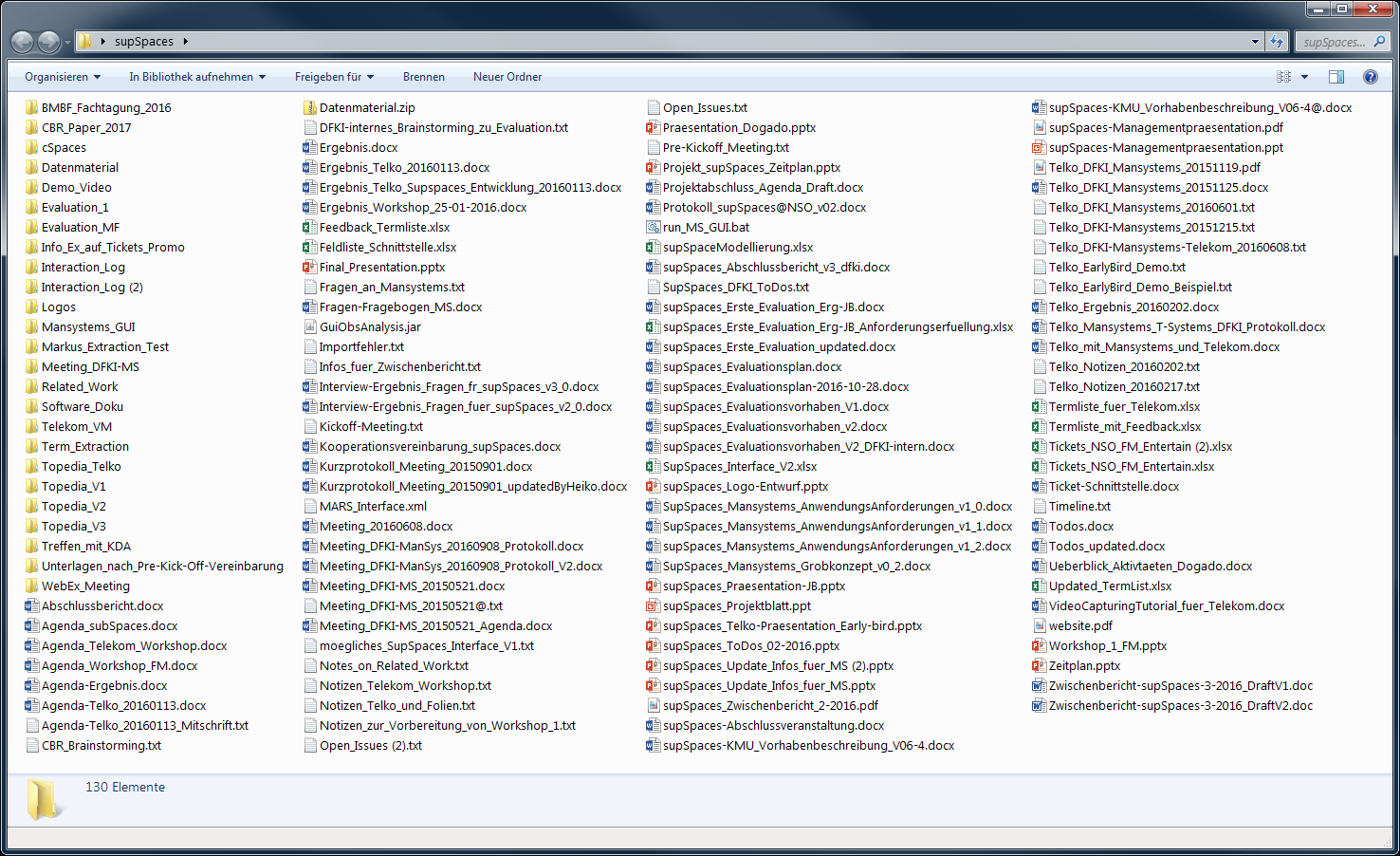}}\\
  \subfloat{\includegraphics[width=1\textwidth]{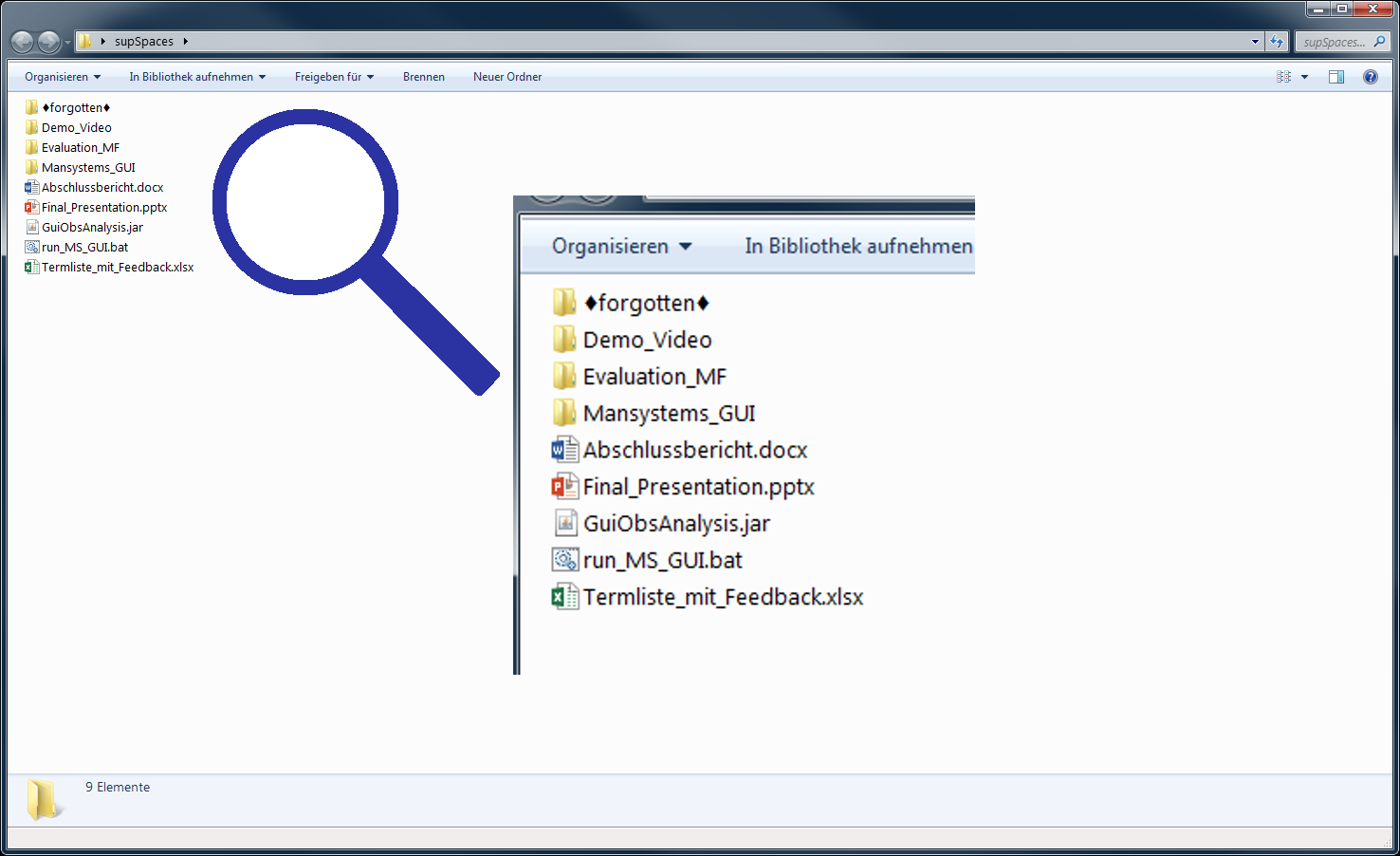}}
  \caption{Example of Reorganization}
  \label{fig:reorg}
\end{figure}

\textbf{Intelligent Folder Injections}.
A last example has already been presented in Figure \ref{fig:contextSwitching}, when user switched from one context to another, the context sidebar as well as all injections (i.e. ``current context'' folder in various applications) were dynamically reorganized to fit the newly selected context's content.
While this can already be seen as a simple form of reorganization, the idea can be further extended.
Using so-called \emph{Intelligent Folder Injections (IFI)}, the current context's content can be further divided with regard to certain criteria.
For example, an IFI could contain less important items of this context that are, however, associated with hot topics of other contexts.
The aforementioned last focus could be another IFI.
Another example, we often use is the binary division of a context's context in an active and a forgotten part, whereas the forgotten part is hidden by default but accessible using another IFI.
This is summarized in Figure \ref{fig:IFIs} depicting a context with three IFIs indicated by black diamond symbols enclosing their name.
Figure \ref{fig:reorg} shows the example of a real-world context, the one of a project not visited for several years.
After returning to that context, users would typically find a situation like in the upper screenshot: an overwhelming number of files and folders and it is at first unclear what were the important items.
In the lower screenshot, items with low MB scores are hidden by default (accessible via a ``forgotten'' IFI) and only the most relevant remain, similar to the calendar entry example in Figure \ref{fig:fadingOut}.
Since screenshots can hardly get across the dynamics of interacting with such a system, we encourage readers to watch a short technical demonstration video available online\footnote{
  \url{https://comem.ai/cspaces/} or \url{https://arxiv.org/src/1805.02181v1/anc/demo_video.mp4}
}.

In the aforementioned multi-month user study, participants unanimously agreed to the statements that IFIs are easy to understand and that they are an easy way to work with an highly autonomous assistant like cSpaces and the results of its hiding, condensation or reorganization measures.
The previous examples apply to temporal as well permanent reorganization.
Further examples of the latter are presented in the next section.

\subsubsection{Permanent Reorganization}

\textbf{Photo Preservation in CoMem}.
The last section showed examples that can be adjusted to work temporally as well as permanently.
Another example of permanent reorganization is related to the \emph{Preservation Value (PV)}.
The PV was designed as a long-term counterpart to Memory Buoyancy.
A first prototypical implementation focusing on photo preservation assessed six dimensions:
\emph{user investment} (e.g. usage or number of annotations or comments added to an item),
\emph{gravity} (e.g. type-based heuristics or connectivity in the KG, in particular closeness to important things),
\emph{social graph} (e.g. relations to important persons, certain persons on a photo),
\emph{popularity} (e.g. ratings, number of views),
\emph{coverage} (e.g. reaching a certain coverage of a user's data, e.g. at least one photo per photo collection) and
\emph{quality} (e.g. photo quality).

Based on a conducted survey on personal preservation of photos \cite{WoltersNivenRunardotter+2015}, we could identify four different personas along two key preservation dimensions: \emph{loss} (worried about losing important photos) and \emph{generations} (importance of preserving important photos for future generations).
For each dimension, two habits could be clustered:
for loss \emph{safety in redundancy ($L_1$)} and \emph{file and forget ($L_2$)}, and for generations \emph{curators ($G_1$)} and \emph{filing first ($G_2$)}.
This led to four different groups: \emph{Safe Curator} ($L_1$/$G_1$), \emph{Safe Filer} ($L_1$/$G_2$), \emph{File \& Forget Curator} ($L_2$/$G_1$) and (filers that) \emph{File \& Forget} ($L_2$/$G_2$).
Next, for each persona, a preservation strategy has been defined, each having different emphases (weights) in the six dimensions mentioned before.
Based on the \emph{Dempster-Shafer Theory of Evidence} \cite{GordonShortliffe1984,YagerLiu2008}\footnote{
  According to this approach, two evidence scores $v$ and $w$ ($v, w \in$ [0,1]) are added as follows: $v \oplus w := 1 - (1 - v) \cdot (1 - w)$. For example, the $\oplus$-sum of 0.6 and 0.7 is 0.88.
}, all indicators of each dimension are summed up and their sum is weighted according to the chosen preservation strategy to yield an information item's PV.
For curators more emphasis is put on investment and gravity, whereas popularity and quality have higher weights for filers.

\begin{figure}
  \centering
  \includegraphics[width=1\columnwidth]{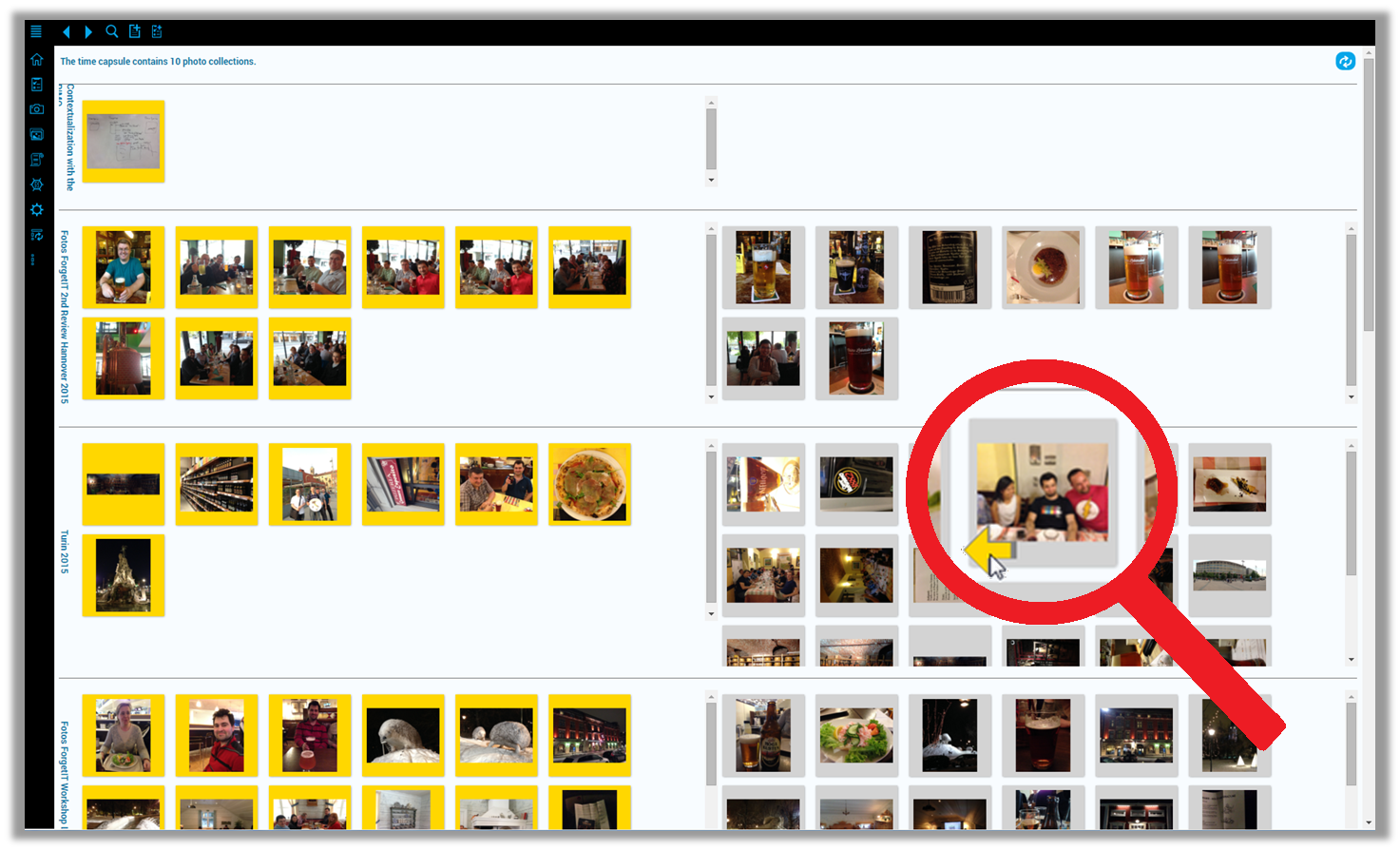}
  \caption{``Time capsule'' feedback interface of photo preservation study}
  \label{fig:photoPreservation}
\end{figure}

To evaluate the PV, a metadata generator was developed, which could be used to create artificial usage profiles based on parameters like number of clicks, length of comments, number of tags, frequency of photo revisitation, etc.
For each aspect, value ranges and simple distributions could be chosen.
By mixing different combinations of usage patterns, we could train and tweak the system with a lot of different user profiles.
Then, this PV calculation was integrated into CoMem and evaluated in a three-week user study involving ten participants.
In the beginning, their PIMO was bootstrapped by the experimenter doing initial interviews and creating relevant persons, locations, topics, etc. for them in CoMem.
Using various photo-related GUIs, each participant could then manage several photo collections having access to CoMem 24/7 as a cloud service.
After three weeks, they were shown a ``time capsule'' feedback interface (depicted in Figure \ref{fig:photoPreservation}), i.e. a split-screen that showed photos to be preserved (high PV) on its left side and the other ones (low PV) on the right.
Participants could move photos from one side to the other to correct the system's decisions.
The system selected between 55\% and 83\% of users' photos to be preserved (mean: 72\%, median: 73\%) and eight of ten participants agreed with 87\% to 96\% of its decisions (mean: 91\%, median: 93\%).
For one participant the agreement was only 76\% and a last one disagreed to such a large extent that using the feedback interface was omitted.

\textbf{General Preservation in CoMem}.
In later studies, preservation was extended to other information items as well combined with a mechanism to move less important items to an archive.
Further details on these studies can be found in \citet{Jilek2023PhD}.

\begin{figure}
  \centering
  \includegraphics[width=1\columnwidth]{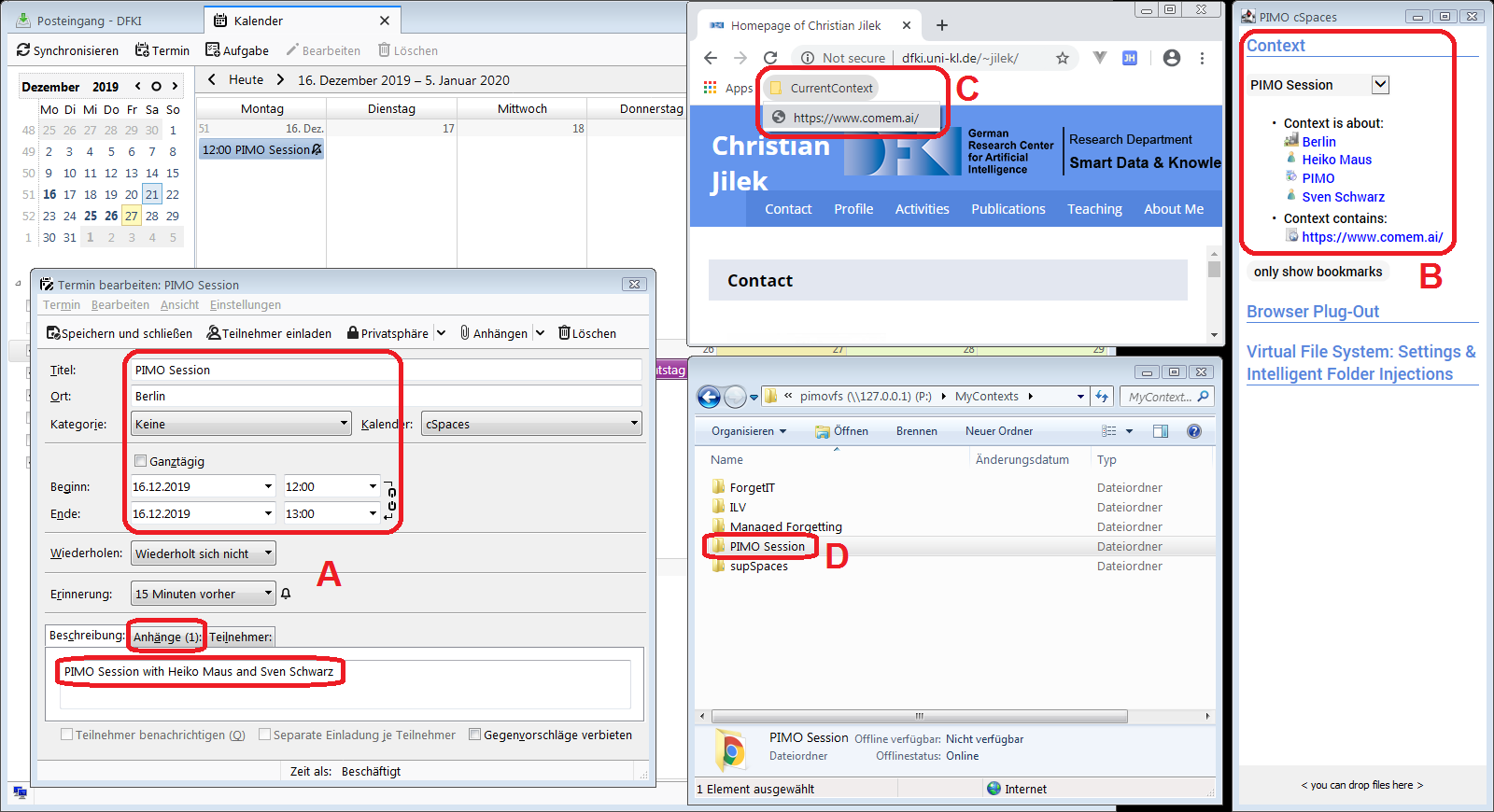}
  \caption{Spawning of a Context Space when creating a new calendar entry}
  \label{fig:calendar2context}
\end{figure}

\textbf{Automated Context Management}.
Another example of permanent reorganization is cSpaces' automated context management.
The idea is to support the full lifecycle of a user context from its initial spawning to splitting and merging over time and forgetting (archiving or deletion) in later phases.
Figure \ref{fig:calendar2context} shows the example of a context space spawning in cSpaces:
A user creates an entry in their calendar (A), which triggers the creation (spawning) of a new context space.
The context can be selected in the sidebar (B), for example.
We also see that information extraction has been performed identifying relevant persons, locations, etc. in the calendar entry, and a mentioned website has already been added to the context.
Being a bookmark, the website is then automatically available in the web browser's ``current context'' folder (C), and the context is also browsable in the file system (D).
Thus, users can directly start working with and in this context space, for example by dropping additional files into it, etc.

Automated context management is a large topic of its own and still under heavy development.
For detailed intermediate results please see \citet{Jilek2023PhD}.

\phantom\\

Due to their autonomous character and the aforementioned paradigm shift, self-organizing and especially forgetting-enabled PKA raise questions with regard to search and trust that are discussed in the following.

\subsection{Searching Forgetting-enabled Information Systems like cSpaces}
In the beginning of this paper, we already mentioned that searching \emph{forgetting-enabled information systems (FIS)} is more challenging than traditional search.
For example, if no or seemingly incomplete search results are shown, users may start to wonder whether they entered the ``right'' keywords or whether they really saved the item they are now looking for.
Being aware that the system is forgetting-enabled, they may also ask themselves whether the searched items have been forgotten by the system and are therefore not showing up.
This may actually be correct because a user is maybe looking for something not used, mentioned or otherwise ``stimulated'' (e.g. by mentioning related topics) for a very long time and now wants to remember or go back to the item.

\begin{figure}
  \centering
  \includegraphics[width=1\columnwidth]{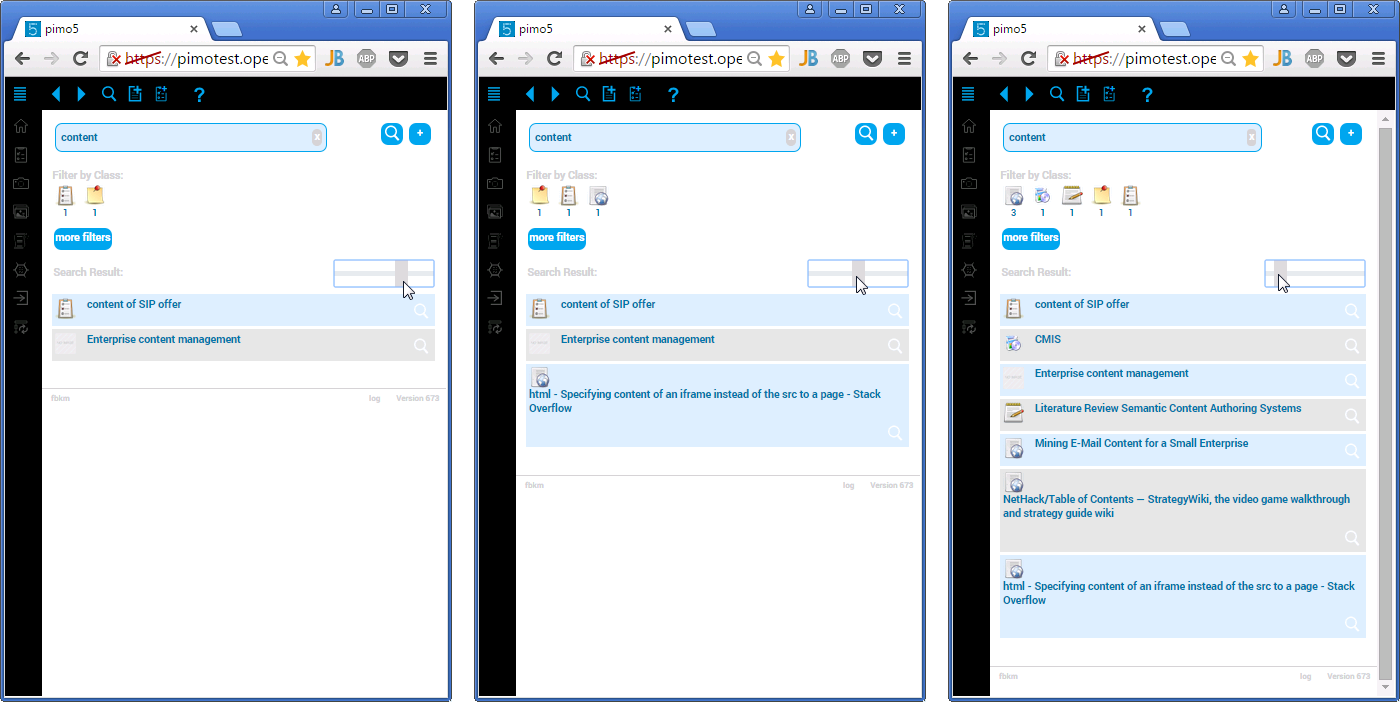}
  \caption{FIS search prototype I: Dragging the Memory Buoyancy slider}
  \label{fig:search1}
\end{figure}

\begin{figure}
  \centering
  \includegraphics[width=1\columnwidth]{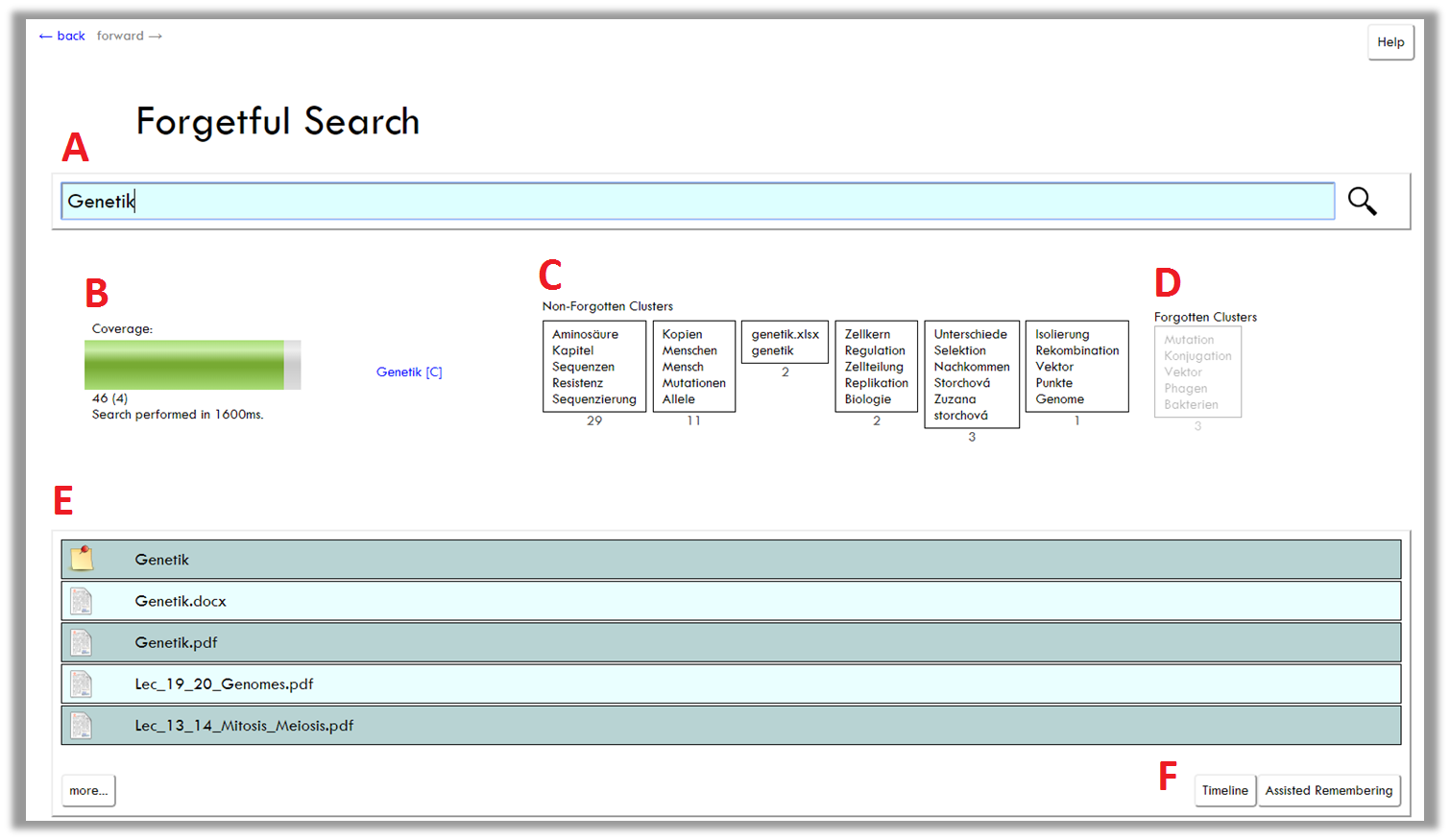}
  \caption{FIS search prototype II: Coverage indication and contextual clustering}
  \label{fig:search2}
\end{figure}

We addressed these issues in several FIS search prototypes, of which two are presented in the following.
Figure \ref{fig:search1} shows our first prototype developed as part of CoMem.
Using a slider, users could decrease the Memory Buoyancy threshold leading to more and more ``forgotten'' items to show up as shown in the three screenshots from left to right.
This led, however, to pathological behavior by users dragging the slider to the left (low MB) instead of rephrasing their query if search results were not satisfactory.
Issues and doubts as described above were still present.
We thus developed a second prototype as shown in Figure \ref{fig:search2}.
Besides the typical input field (A) and the list of results (E), there is a coverage indicator (B) showing how many found items are in the ``active'' part of the system and how many in the ``forgotten part''.
Users thus get an idea of what actually is or was there without being additionally flooded by possibly rightfully forgotten items.
In addition, contextual clustering is performed, showing topic clusters for active (C) and forgotten parts (D) of the user's information sphere allowing for further drill-down of the search.
More details and evaluation results can be found in \citet{Jilek2023PhD}.

\subsection{Trust in Highly Autonomous Assistants like cSpaces}
In scenarios involving a highly autonomous PKA like cSpaces that reorganizes a person's information sphere, trust is particularly important.
This comprises trust in search results (see last section) and the system's actions and not losing the feeling of control.

Together with colleagues and students of psychology/ergonomics, we conducted an online study, in which 140 participants assumed the role of a knowledge worker being supported by cSpaces in the form of a sidebar on one side of the screen.
In three different situations, participants were shown three different variants of status messages of cSpaces.
These messages were phrased to correspond to levels of the \emph{Situation Awareness-based Agent Transparency (SAT) model} by \citet{ChenProcciBoyce+2014}.
A first variant only stated \emph{goals and actions} (e.g. a new context has been created by cSpaces).
A second variant additionally explained \emph{reasoning} behind the action (e.g. the last four actions strongly addressed a certain project and therefore the context has been created).
A third variant also gave \emph{projections} on the assistant's future actions (e.g. if additional evidences confirmed the context switch, the user's last three actions would also be assigned to the newly created context).
Findings of the study were as follows:
\begin{itemize}
\item Trust and intention to use increased as a function of higher transparency.
\item There was no moderation effect of propensity to trust.
\item Transparency perceptions differed across situations.
\item Participants formed a preference.
\end{itemize}
In conclusion, one can say that more explanations were not necessarily perceived as more transparency, and for different situations different levels were preferred.
An assistant's explanations should thus be adaptable and their amount adjustable.
For example, in the beginning, when working with such a system is still a new experience for users, more explanations are presumably helpful.
After some time, when users are more accustomed to the system, they may prefer to regulate down their amount.
For further details please see \citet{Jilek2023PhD} or a paper about the experiment that is currently in preparation\footnote{
  A.-S. Ulfert-Blank, J. Knabe and C. Jilek: \emph{Forgetting-enabled AI systems: Exploring the role of transparency perceptions and trust} (paper in preparation)
}.

\phantom\\

Having presented several experiments and results with regard to knowledge graph construction and personal knowledge assistants, this paper's main part is concluded by a section on related industry use cases.
\section{Industry Use Cases}
\label{sec:Industry}

The German Research Center for AI (DFKI) is application-oriented and one of its missions is transfer to industry. With our research on corporate memories and assistance of information and knowledge workers presented in this article, we could fulfill this mission and realize several industry use cases which we will address in the following sections.

\subsection{The Corporate Memory CoMem at enviaM} \label{sec:comem_enviam}

The DFKI cooperates with the German energy service provider envia Mitteldeutsche Energie AG (enviaM)\footnote{\url{https://www.enviam-gruppe.de/}}, a subsidiary of E.ON SE, since 2017 to demonstrate the practicability of the Corporate Memory approach and transferring the CoMem infrastructure into productive use.  
As a first pilot, CoMem was implemented in the real estate services for property management to demonstrate its feasibility\footnote{This scenario is explained in more detail at \url{https://comem.ai/home/showcase}}. Meanwhile the scenario is in productive use together with further domains such as municipal support, compliance management, construction projects, contact search and company-wide inbox management. Further research is in progress to support knowledge workers in accounting (see, e.g., \cite{SchulzePelzerSchroeder2022,BenndorfMausStein+2022}).

The challenge in the real estate department was a high load on the knowledge worker for assembling all relevant information from distributed and heterogeneous sources for a specific task. 
Employees need a wide variety of sources to do their daily work: a network drive that has grown over several years as a document storage location; a legacy system for the maintenance of properties, real estate parcels and ownership of several corporate clients; various data collections (from other departments), preferably in Excel format, with specialized information that is needed repeatedly for different tasks, such as asset values from the enterprise resource planning system (ERP); standard office applications, such as e-mail and the intranet. 
Each of these sources uses different terminology, making it more difficult to identify information about, say, a specific parcel of land.
As a result, working with this collection of data silos is challenging and sometimes tedious, especially when cross-silo information is needed, and even more so if it crosses departmental boundaries.

With the heterogeneous file and document collections, databases and legacy systems for dedicated services, the scenario we face is not unlike that of many other organisations with grown infrastructures. In particular, legacy systems - although adequate for their intended purpose, the data is inaccessible and not easily available outside these systems for various other use cases. 
For a variety of reasons, such an infrastructure setting is not on the agenda to be replaced or unified by a comprehensive restructuring (for instance, as they are not part of primary processes, potentially too high complexity for a single system, no tailored solution on the market, supposedly high investment).

\begin{figure}
  \centering
  \includegraphics[width=1\columnwidth]{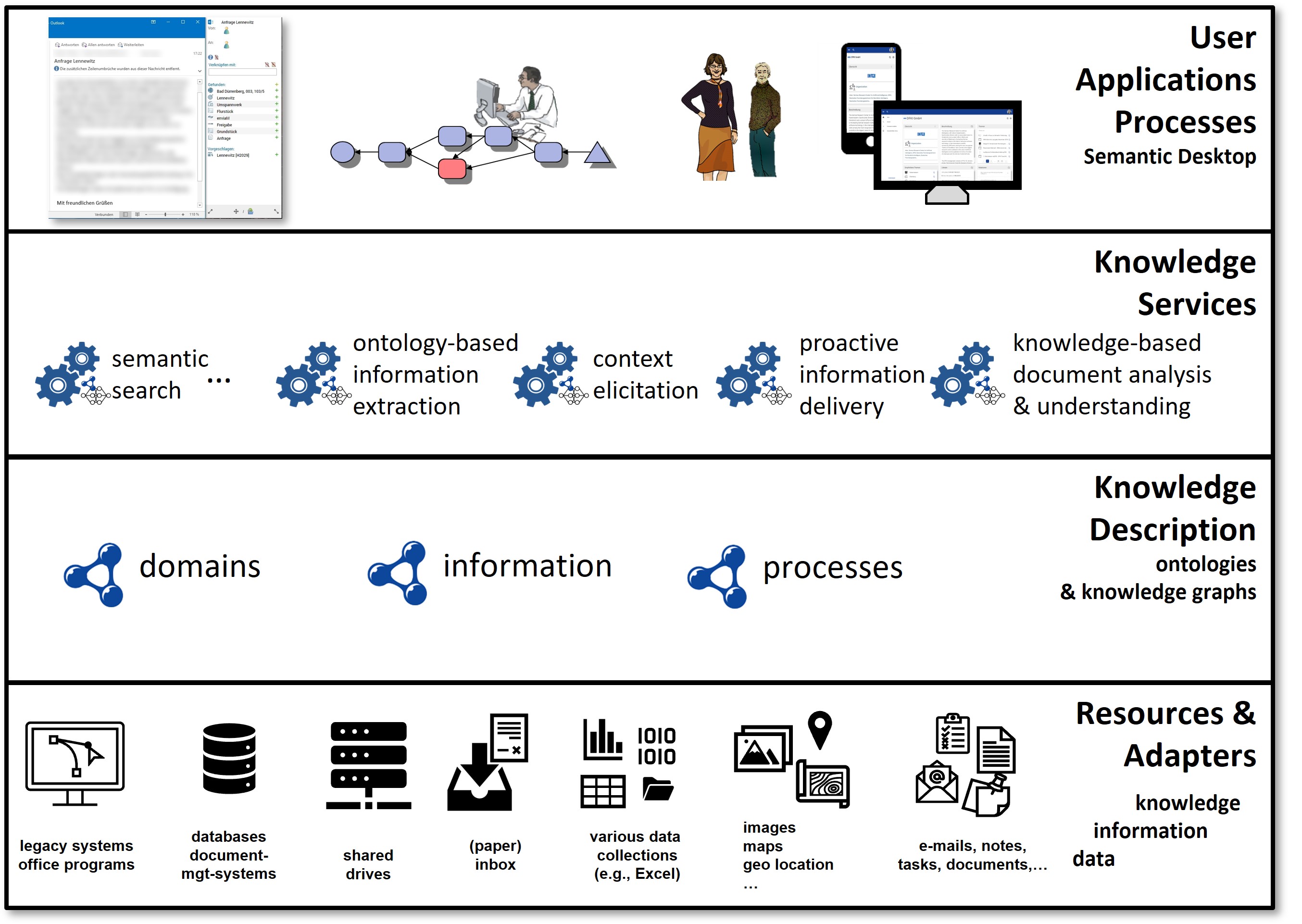}
  \captionsetup{format=hang}
  \caption{The layers of the Corporate Memory CoMem (based on \citet{AbeckerBernardiHinkelmann+98}; adapted from \citet{RissMausJavaid+2020})}
  \label{fig:omlayers}
\end{figure}

To allow for personalized assistance in this corporate environment, we tackled the scenario by the means of the CoMem infrastructure along the four layers for Corporate Memories depicted in \Cref{fig:omlayers} which is based on the work of \citet{AbeckerBernardiHinkelmann+98} from our department.
Let us investigate these four layers to explain our approach: first, in the \emph{resources \& adapters layer} the variety of sources, such as Excel files, shared drives, document and image collections, geo-information and the real estate information system, has been connected by dedicated adapters transforming the respective data structure to the models defined by the ontologies in the knowledge description layer and populating knowledge graphs accordingly.
Depending on the source, different methods were applied, e.g. constructing KG from spreadsheets as detailed in \Cref{sec:KGC_corporate}. 
Initially, data dumps from legacy systems were used for rapid prototyping in the pilot phase. In the meantime, the data sources of enviaM are connected via the industry-standard enterprise middleware \emph{inubit}\footnote{\url{https://www.virtimo.de/en/products-en/\#inubit}} providing an API and adapters to connect to enviaM's heterogeneous infrastructure. 
We defined one central \emph{CoMem-inubit} component between CoMem and inubit, realizing a designated API used by CoMem to access each connected system in a homogeneous way via this one central access point.
This API specifically contains all required methods to enumerate attached systems, subsets and data objects as well as ways to crawl, get, update and check data.
That way, the intra-company technical access to the legacy systems, user-management, privileges etc. can be treated and developed by the IT personnel of the company, and CoMem just needs access to this one central component.
CoMem-inubit provides the metadata of the objects (in attached systems) including provenance data of origin and timestamps in a well-defined metadata format in XML.
The metadata for each data object also contains permission information which users and groups in the company are allowed to access this specific object.
This information is processed and expressed in the resulting knowledge graph to guarantee that users can only see what is disclosed to them.
For this, the permissions of legacy systems are taken into account. Additionally, administrators of each system can enable explicit access to a larger audience.
Thus, CoMem respects the underlying permission rules while still allowing broader access where it is needed to bridge pure-technical barriers.

\begin{figure}
  \centering
  \includegraphics[width=1\columnwidth]{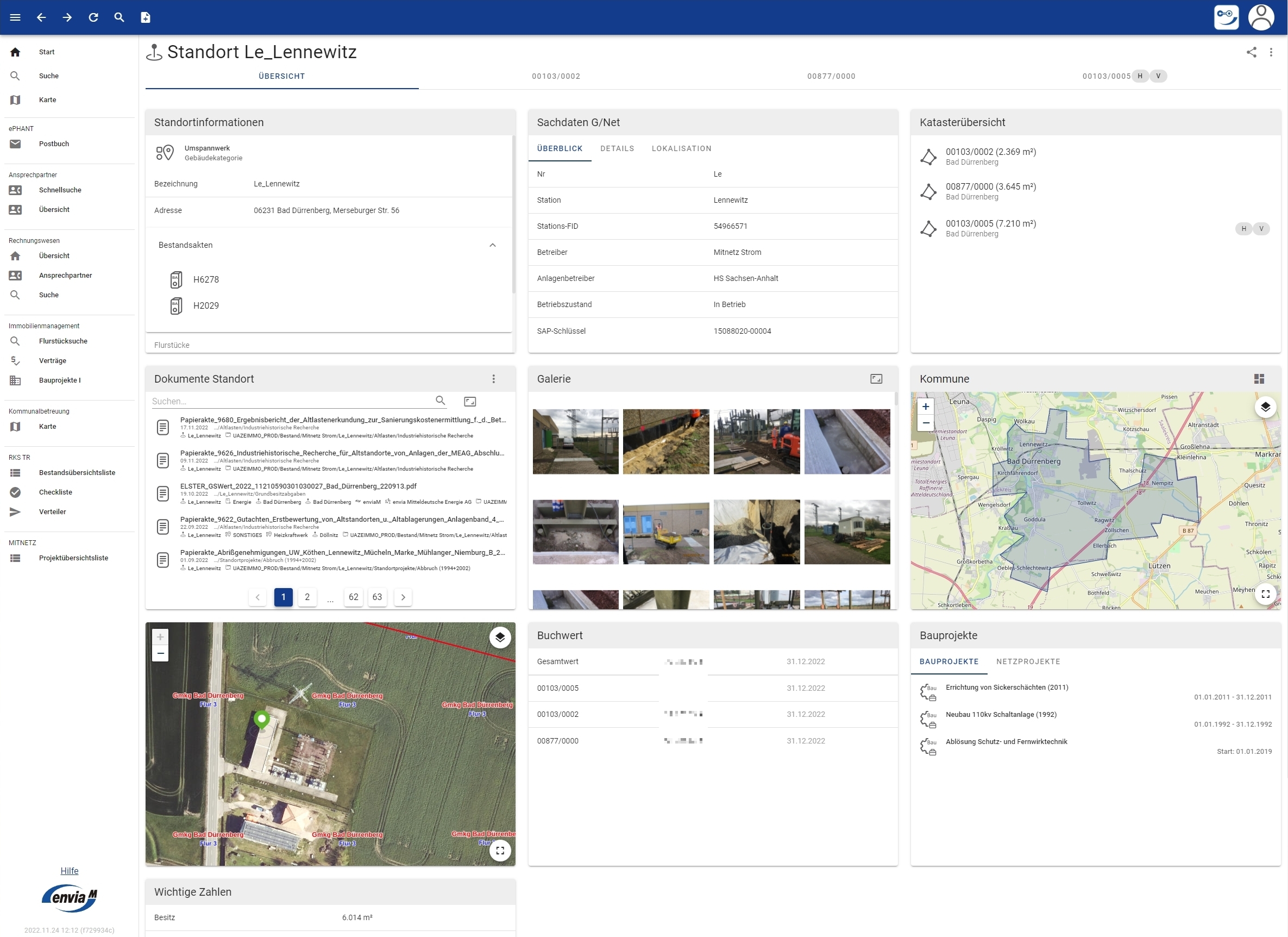}
  \captionsetup{format=hang}
  \caption{CoMem dashboard for the site of a power station providing a view on the knowledge graph derived from various external sources.}
  \label{fig:ilvDashboard}
\end{figure}

The \emph{knowledge description layer} consists of ontologies and knowledge graphs accessible via an API allowing to create, manipulate and search. For each domain,  dedicated ontologies are designed which enable expressiveness for use cases such as supporting purchase-to-pay processes \cite{TrkmanMcCormack2010} in industry as detailed in \citet{SchulzeSchroederJilek+2021}\footnote{The ontology is available at \url{https://purl.org/p2p-o}}.
The ontologies reside in an ontology server allowing to model, version and query the ontologies required for the respective domain and use cases. The KGs are stored in the CoMem knowledge repository which is a relational database with a triple store layer on top, full-text and concept indexing capabilities and access rights on the knowledge graphs.
Thus, in the real estate scenario, the knowledge repository stores statements about sites, parcels, locations or topics; resources such as contracts, e-mails or official letters including statements such as all KG concepts mentioned in a document derived from content analysis and metadata such as provenance, access rights or memory buoyancy. 
It enables CoMem to interconnect the sites and cadastral information from the real estate information system, book values from the ERP, with documents and images as depicted in \Cref{fig:ilvDashboard}. 
As the ontology for the real estate domain provides a basic understanding on a site and geo-location etc., it was extended with further domains from different departments such as municipal support and construction projects. This enables users from different departments to incorporate this information -- previously inaccessible or time-consuming to compile -- into their work context. 
As of 2023, the enviaM knowledge graph consists of 9 million things (mainly documents) and approx. 200 million triples.  

The knowledge description layer allows to store, access, manipulate, interconnect and reason over information regardless of the original data source and format. This enables the  \emph{knowledge service layer} to abstract from the data and sources and establish its services (mainly) on the knowledge graphs and referenced resources.
Here, services such as expectation-driven and process-oriented document analysis and understanding, context-elicitation and proactive-information delivery are integrated (see, e.g., \citet{AbeckerBernardiMaus+00}).
For enviaM, we use our specialized NER approach \cite{JilekSchroederNovik+2019} to analyze all textual resources to enrich the KG with statements of concept occurrences which is used in the import as mentioned above, as well as in our end-user search, which combines full-text and concept-based search.

\begin{figure}
  \centering
  \includegraphics[width=1\columnwidth]{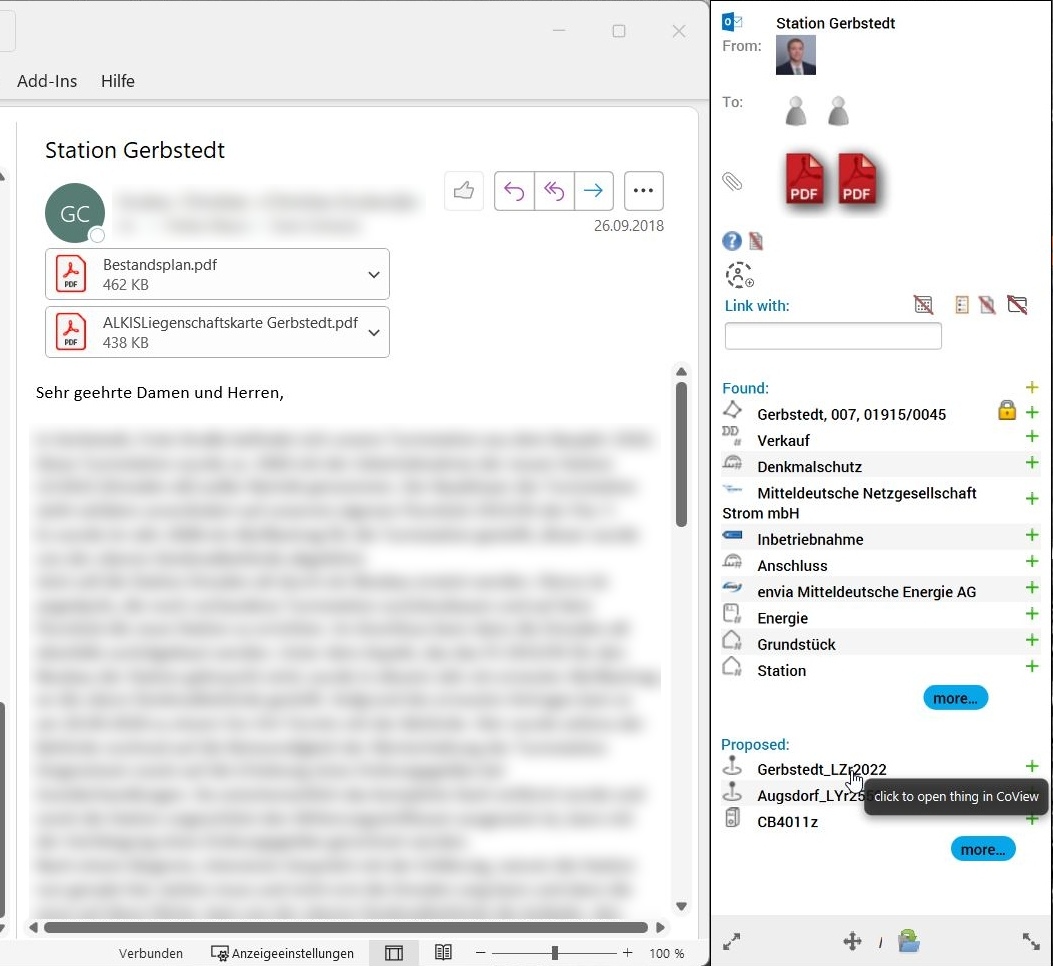}
  \captionsetup{format=hang}
  \caption{The Semantic Desktop enables pro-active information delivery for an lengthy e-mail in the real estate scenario.}
  \label{fig:pid_enviam}
\end{figure}

Finally, in the \emph{application layer}, various interfaces such as the CoMem CoView -- a web browser interface (see ~\Cref{fig:ilvDashboard}) -- give access to the knowledge repository and integrating knowledge services such as dashboards, search or a semantic editor \cite{EldesoukyBakryMaus+16}. 
As detailed in the previous sections, our mission is to reduce the effort for individual knowledge workers and to embed the assistance into their daily work to derive context and realize immediate benefits. The Semantic Desktop is therefore one of the pillars in the application layer integrating users on their desktops.
At enviaM, we started with providing a lightweight version of the sidebar detailed in \Cref{sec:userInterfaces} for assisting in e-mail scenarios as depicted in \Cref{fig:pid_enviam}.
The Microsoft Outlook plug-out enables the CoMem Semantic Desktop client to get the event that the user clicked the e-mail and access to its content.
Using our NER and the domain-specific knowledge graphs, content analysis on the e-mail then detects known entities such as topics, processes or the specific parcel of land. This is shown in the sidebar as a kind of specialized summary of the text using enviaM terminology and internal entities (heading \emph{found}). 
These extracted concepts serve as a context for the pro-active information delivery, here, a query on the KG resulting in a list of relevant concepts. For the depicted e-mail, this results in proposing adequate sites (heading \emph{proposed}). Clicking on the concept opens the dashboard of the respective site (see \Cref{fig:ilvDashboard}) providing the user with immediate access to a range of information required for the task of the e-mail reducing the tedious effort of consulting several data sources.

With the CoMem system being in productive use at enviaM and spreading to further departments, it provides us with insights on the feasibility of the approach as well as further research challenges to pursue, such as the previously mentioned assistance in purchase-to-pay processes by \citet{SchulzePelzerSchroeder2022} or enhancing the assistance with robotic process automation as introduced in \citet{ZeyenKochSchwarz+2022}.

\subsection{supSpaces -- cSpaces in IT Support Scenarios}
The term \emph{supSpaces} is short for \emph{support (knowledge) spaces for knowledge management in IT support}.
It was a two-year project together with Mansystems (now CLEVR\footnote{
  \url{https://www.clevr.com/}
}), a developer of service management applications, DFKI and two application partners, Deutsche Telekom\footnote{
  \url{https://www.telekom.de/}
}, a German telecommunications company, and Dogado\footnote{
  \url{https://www.dogado.de/}
}, a German cloud service provider.
In the scenario, a context space is spawned and develops ``around'' a service ticket covering an incident reported by either a monitoring software or by a customer.
By claiming a ticket, a clerk creates and enters such a supSpace.

\begin{figure}
  \centering
  \includegraphics[width=1\columnwidth]{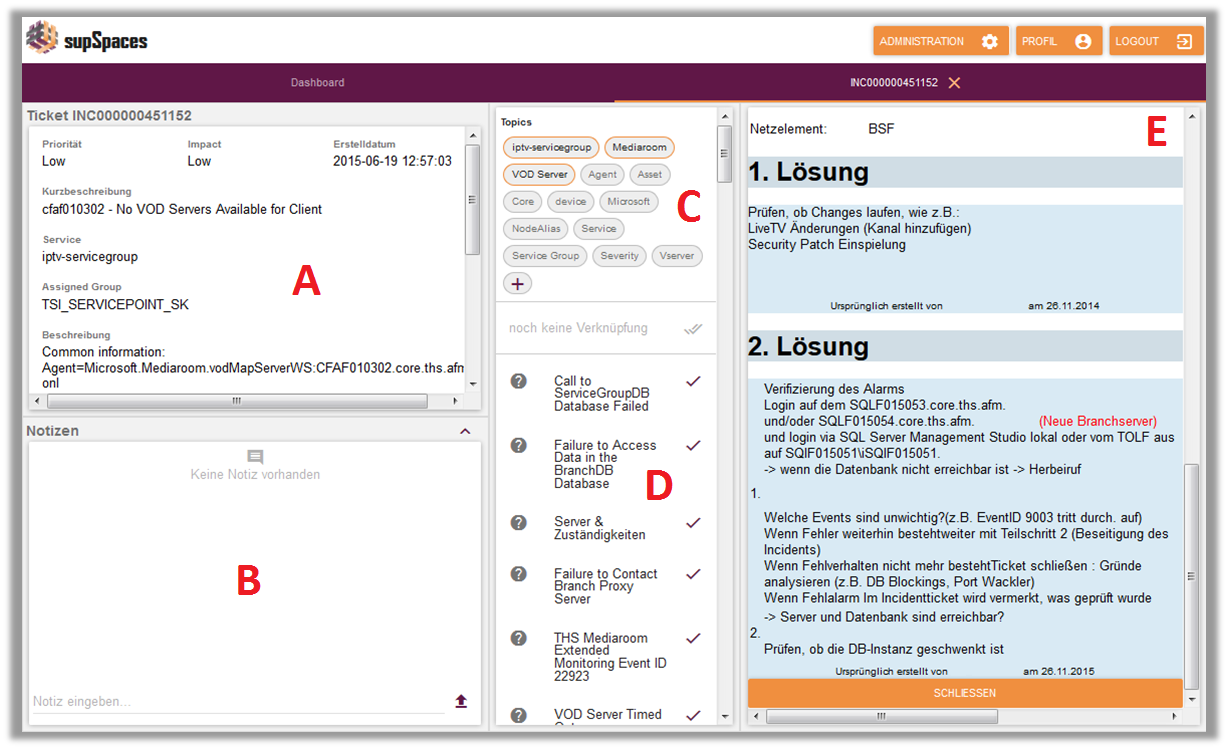}
  \captionsetup{format=hang}
  \caption{supSpaces dashboard}
  \label{fig:supSpaces}
\end{figure}

Its graphical user interface is a dashboard depicted in Figure \ref{fig:supSpaces}.
Highlighting A shows the claimed ticket.
There is also a text area for taking notes in that particular context space (B).
After the initialization of the context space, the system first performs information extraction on the ticket identifying entities of the knowledge graph in the text as shown in the the figure's middle section (C).
Based on the ticket analysis, support tickets of the past (as well as their solutions, if available) and relevant resources are recommended (D).
Clicking a resource loads its content inside the dashboard (E).
Following CoMem's idea of being a meta-system ``on top'' of legacy systems, clerks may also follow a link to jump to the original location of the resource.
Later versions also allowed for keyword-based search.
By confirming relevant topics and helpful material, writing comments, associating items with the ticket's context, etc., clerks may easily document their progress, solution attempts (particularly helpful if the ticket cannot be solved before a shift change) and finally their solution.
During the processing of the ticket, the supSpaces system live-updates its list of proposed supportive material after each captured user activity -- one of the reasons why we needed the aforementioned real-time-capable information extraction approach.
Then, the system finally saves the supSpace, i.e. the ticket together with its rich context consisting of all information described before.
If such a problem or a similar one occurs (again), the system will bring up the previously captured supSpace.
Users can then decide whether the solution to the old problem can actually be transferred to the new one, which is now considerably easier given the old ticket's additional contextual information.
The approach was mainly realized by means of \emph{Case-Based Reasoning (CBR)} \cite{AamodtPlaza1994}.
The supSpaces prototype as described above was evaluated in an expert walkthrough (structured interview).
A clerk of Deutsche Telekom used the system to solve a set of past tickets and rated the system's different features while doing so.
The overall feedback was generally very positive.
The idea of a supSpaces prototype was first mentioned in \citet{JilekSchwarzMaus+2016} and a retrospective overview (of which this is an excerpt) was provided in \citet{Jilek2023PhD}.

\subsection{Wacom Digital/Semantic Ink}
The aforementioned real-time-capable and also inflection-tolerant information extraction approach, together with the idea of CoMem as a whole, was the main focus in a collaboration with Wacom\footnote{
  \url{https://www.wacom.com/}
} that started in 2018\footnote{
  \url{https://www.wacom.com/en-us/about-wacom/news-and-events/2018/1335}
}.
DFKI served as a consultant to help Wacom develop tailored versions for their \emph{Semantic Ink} technology which enhances their Digital Ink with knowledge graphs, in particular \emph{Wacom Notes}\footnote{%
  \url{https://www.wacom.com/en-en/products/wacom-notes}
}, a worldwide available product.

\subsection{Cybermapping of the Financial System}

Ongoing digitization and outsourcing of processes increase the degree of networking between financial systems and IT service providers.
Cyber incidents can significantly impair the provision of important functions of the financial system and thus pose a risk to financial stability. 
To identify these vulnerabilities, it is important to understand the relationships between the financial network and the cyber network which can be realized with a cybermapping \cite{brauchle2020cyber}.
The \emph{International Monetary Fund (IMF)} and the \emph{European Systemic Risk Board (ESRB)} assess cybermapping as an important part of cyber risk monitoring, yet point out a current lack of data \cite{adelmann2020cyber}.
Therefore, together with the \emph{Deutsche Bundesbank [German Federal Bank]} we are building such a model in a transfer lab\footnote{\url{https://www.dfki.de/en/web/news/cybermapping-and-ai}} by utilizing knowledge graphs and semantic technologies.
The aim of this project is the development of an approach that enables the analysis of enterprise networking, cyber risk concentrations and transmission of cyber shocks.
This requires methods for collecting, linking, compiling, maintaining, analyzing, and evaluating data sources necessary for identifying systemic cyber risks.

\subsection{Knowledge Graph Tooling}

Once we deploy knowledge graphs, we provide tools for accessing and working with them for our industry partners.
Our customers are often not aware of fundamental topics about Semantic Web and they rather seldom use such technologies.
To still enable their software developers to take part in our projects, we support them with appropriate Application Programming Interfaces (APIs).

\textbf{SPARQL REST API}.
One is a REST API that mediates to SPARQL \cite{w3cSparql2008,w3cSparql2013} which is presented in \citet{DBLP:conf/esws/SchroderHBEKS18}.
By using the path metaphor in the URL request and turning them to SPARQL queries, the API lets users navigate through the KG.
Additionally, graph representations are bidirectionally transformed into an JSON-based \cite{rfc4627,rfc8259} object view for an easy data comprehension.
Usual \emph{Create, Read, Update and Delete (CRUD)} operations are supported with typical POST, GET, PATCH and DELETE requests.
More expressive features are provided such as asterisk wildcards in the path, the use of property paths \cite{w3cSparql2013} and RQL\footnote{\url{https://github.com/persvr/rql}}.

\textbf{Linked Data Application Framework}.
To ease the development of linked data applications and enable rapid prototyping for software engineers, we designed a dedicated framework for them in \citet{SchroederJilekDengel2021ldaf}.
Here, the idea of an API for developers is complemented with an HTML representation for non-technical users and RDF documents for Semantic Web practitioners.
Moreover, the framework features an authentication mechanism, linked data resources, bidirectional RDF/JSON conversion and HTML template rendering.

\textbf{RDF2RDB REST API}.
Another way to access KGs is a conversion of them to a data structure the developers are familiar with, for instance a \emph{Relational Databases (RDB)} \cite{rdb-history} which is shown in \citet{SchroederSchulzeJilek+2021}.
Here, the RDF model of a KG is automatically turned into a database in a type-store fashion \cite{DBLP:journals/ker/MaCY16}.
Additionally, REST API code is generated for an easy access, similar to the previous approaches.
Using datasets from \emph{Linked Open Data (LOD)} \cite{BernersLee2006}\footnote{\url{https://lod-cloud.net/}} and a benchmark \cite{DBLP:books/igi/11/BizerS11}, we showed that our approach generates a meaningful number of tables by inferring cardinalities from RDF data.

\textbf{Simple RDFS Editor}.
Working with KGs often requires the definition of suitable ontologies describing certain domains.
While ontology editors like Protégé \cite{DBLP:journals/aimatters/Musen15} are powerful, they usually pose a steep learning curve for beginners. 
Therefore, we designed an RDFS ontology editor\footnote{\url{https://github.com/mschroeder-github/simple-rdfs-editor}} with a simplified feature set and convenient operations.
An online version lets multiple editors work collaboratively on one ontology, similar to WebProtégé\footnote{\url{https://webprotege.stanford.edu/}}.

\textbf{Hephaistos Toolkit}.
All mentioned approaches and tools related to KG construction (see Section~\ref{sec:KGC}) are collected in a toolkit called \emph{Hephaistos}\footnote{\url{https://www.dfki.uni-kl.de/~mschroeder/hephaistos/}}.

\section{Conclusions}
\label{sec:Conclusions}
In this paper, we gave a retrospective overview of a decade of research in our department towards self-organizing personal knowledge assistants in evolving corporate memories.
It was typically inspired by real-world problems and often conducted in interdisciplinary collaborations with research and industry partners.
We summarized past experiments and results comprising topics like various ways of knowledge graph construction in corporate and personal settings, \emph{Managed Forgetting} and \emph{(Self-organizing) Context Spaces} as a novel approach to PIM and knowledge work support.
Past results were complemented by an overview on related work and our latest findings not published so far, like a multi-month user study, fNIRS and trust studies.
Last, we gave an overview of our related industry use cases including a detailed look into CoMem, a Corporate Memory based on our presented research already in productive use and providing challenges for further research.
Many contributions are only first steps in new directions with still a lot of untapped potential, especially with regard to further increasing the automation in PIM and knowledge work support.

One of the very recent topics we are currently working on is utilizing methods of deep learning, which is a challenge in our sparse, privacy-sensitive, near-real-time user support scenarios involving highly evolving knowledge graphs.

\subsubsection*{Acknowledgements}
This work was partly funded by the German Federal Ministry of Education and Research in the project \href{https://www.comem.ai/sensai/}{SensAI} (grant no. 01IW20007) and the German Research Foundation (DFG) in the project \href{http://www.spp1921.de:8442/projekte/p4.html.en}{Managed Forgetting} (grant no. DE 420/19-2).

The authors would also like to thank all members of our team of researchers and software engineers as well as our research and industry partners, students and study participants, who supported our work over the years.

\printbibliography

\end{document}